\PassOptionsToPackage{hyphens}{url}
\documentclass[sigconf,nonacm]{acmart}

\usepackage{booktabs}
\usepackage{array}

\AtBeginDocument{\expandafter\def\expandafter\UrlBreaks\expandafter{\UrlBreaks
  \do\a\do\b\do\c\do\d\do\e\do\f\do\g\do\h\do\i\do\j\do\k\do\l\do\m
  \do\n\do\o\do\p\do\q\do\r\do\s\do\t\do\u\do\v\do\w\do\x\do\y\do\z
  \do\A\do\B\do\C\do\D\do\E\do\F\do\G\do\H\do\I\do\J\do\K\do\L\do\M
  \do\N\do\O\do\P\do\Q\do\R\do\S\do\T\do\U\do\V\do\W\do\X\do\Y\do\Z
  \do\0\do\1\do\2\do\3\do\4\do\5\do\6\do\7\do\8\do\9}}

\newcommand{\code}[1]{\texttt{#1}}
\newcolumntype{L}[1]{>{\raggedright\arraybackslash}p{#1}}
\newcolumntype{R}[1]{>{\raggedleft\arraybackslash}p{#1}}
\title[How Much Effort Does an AI Code Reviewer Need?]{How Much Effort Does an AI Code Reviewer Need? Evidence from Claude 5.5 Models on Plausible-but-Wrong Patches}

\author{Zax Shen}
\orcid{0009-0004-8681-3359}
\email{zax@trustmy.bot}
\affiliation{%
  \institution{Trust My Bot}
  \city{Los Angeles}
  \state{California}
  \country{USA}}

\author{Hanling Zhang}
\authornote{Hanling Zhang and Zixu Wang contributed equally.}
\orcid{0009-0003-2681-6614}
\email{cciezhl@gmail.com}
\affiliation{%
  \institution{Trust My Bot}
  \city{Hong Kong}
  \country{Hong Kong}}

\author{Zixu Wang}
\authornotemark[1]
\orcid{0009-0005-1001-8145}
\email{wangzixu0810@gmail.com}
\affiliation{%
  \institution{Independent Researcher}
  \city{Orlando}
  \state{Florida}
  \country{USA}}

\begin{document}

\begin{abstract}
AI coding agents write code faster than people can review it, so teams add AI reviewers. Reasoning models let such a reviewer trade cost for diligence through an effort setting. We asked how much effort a reviewer needs to catch plausible-but-wrong patches, which pass a benchmark's tests but fail stronger ones. Claude Opus 5.5 (preregistered), Sonnet 5.5 and Haiku 5.5 (exploratory) each reviewed the same 90 agent-written Django patches from SWE-bench Verified at low, medium and high effort. By strengthened SWE-ABS tests, 55 patches were wrong and 35 correct. In every model, high effort accepted fewer wrong patches than low (false success .45 against .67 for Opus, .49 against .65 for Sonnet, .51 against .78 for Haiku), a supported difference for Opus and Haiku. For Opus, medium was non-inferior to high within 15 percentage points (difference .018, 95\% CI [$-$.065, .108]) and cost 20\% less, short of the preregistered 30\%. For Sonnet and Haiku, medium cost 39\% and 51\% less than high but was not shown non-inferior. Opus at medium or high agreed best with the labels; Haiku at high cost about 8\% as much per review as Opus at medium, accuracy difference unresolved. Exploratory coding showed that most rejections named the tested defect and that effort added such rejections. No setting is a strong gate: each accepted at least 45\% of wrong patches. For 32 of the 55 wrong patches the issue may not require what the failing test checks; without them, neither accuracy hypothesis holds for any model.
\end{abstract}



\maketitle

\section{Introduction}\label{sec:intro}

Coding agents now produce code much faster than people do. In 806 open-source projects that adopted the Cursor agent, lines of code added rose by 281\% in the first month, though the gain faded after two months \cite{he2026cursor}; in one project, a developer submitted as many agent-written pull requests in three days as they had submitted without an agent in the previous three years \cite{li2025teammates}. Review has not kept pace. In telemetry from more than 10,000 developers, the time spent reviewing each pull request rose 91\% in teams with high AI adoption \cite{farosai2025paradox}, and a study of agent-written pull requests reports a growing imbalance between how fast they are submitted and how fast they can be reviewed \cite{li2025teammates}. Review is becoming the bottleneck, and a common response is a second model that reviews each patch before merge. Trust My Bot (\url{https://github.com/trustmybot/tmb}), an agent harness, gates every push this way. Such a gate is worth its cost only if it rejects broken patches without rejecting good ones.

Reasoning models expose an \textit{effort} setting that controls how much work the model does before it answers. Claude Opus 5.5 offers five levels, with medium as the default \cite{anthropic2026effort,anthropic2026model}. Effort shapes all output, thinking, tool calls and text, and is not a strict token budget \cite{anthropic2026effort}. More effort costs more; whether it also gives better verdicts sets the price of a review gate. The answer may differ by model: a smaller model costs less per review and may need more effort to reach the same verdicts.

Measuring a reviewer needs a reference standard. SWE-bench Verified pairs GitHub issues with tests that people screened as fair \cite{chowdhury2024verified,jimenez2024swebench}, but those tests are weak: SWE-ABS strengthens them \cite{yu2026sweabs}. Its paper reports that the stronger tests reject 19.78\% of the passing patches of its top 30 agents \cite{yu2026sweabs}, and by our count its released verdicts reject 18.6\% of the passing patches of all 39 agents (2,682 of 14,446) \cite{openagenteval2026sweabs}. A patch that passes the Verified tests but fails SWE-ABS is \textit{plausible but wrong}. The benchmark counts it as solved, and a review gate exists to catch it.

We report a paired experiment in which three Claude 5.5 models, Opus, Sonnet and Haiku, reviewed the same 90 patches at low, medium and high effort. The Opus study was preregistered and is confirmatory; the Sonnet and Haiku studies are exploratory. The reviewer setup was fixed; only effort and the model varied. This paper is the first in a series of studies of AI code review; later studies vary a note from the patch's author, the reviewer's skills and voting among several reviewers. We asked:

\begin{itemize}
\item \textbf{RQ1.} In each model, does effort change how often the reviewer accepts wrong patches (false success) and rejects correct ones (false alarm)?
\item \textbf{RQ2.} In each model, is medium effort as good as high on false success, and materially cheaper?
\item \textbf{RQ3.} In each model, what does each step up in effort buy in agreement with the labels, and at what cost?
\item \textbf{RQ4.} How far do strict tests and the model's memory of the upstream fix qualify the answers?
\item \textbf{RQ5.} Across the three models and three efforts, which setting gives the best review gate per dollar?
\item \textbf{RQ6.} When a reviewer rejected a wrong patch, did it name the defect the failing test checks, and what did its findings and false alarms claim?
\end{itemize}

For Opus we preregistered three hypotheses. H1: high effort has lower false success than low. H2, the primary hypothesis: false success at medium is at most 15 percentage points above high. H3: geometric-mean cost per review at medium is at least 30\% below high. We ran the same three hypotheses on Sonnet and Haiku after their data were in, so those tests are exploratory. RQ3 to RQ6 are exploratory for every model. Every table and figure shows all three models and marks Opus as preregistered and Sonnet and Haiku as exploratory, except the recall analysis, which was run for Opus only. Every rate measures agreement with labels derived from the official fix (Section~\ref{sec:labels}); RQ4 asks how much that matters.

The paper makes five contributions:

\begin{enumerate}
\item A paired experiment of 810 reviews in sealed containers, 270 per model, of which the Opus part was preregistered, with public data, analysis code and spend record.
\item Evidence, confirmatory for Opus and exploratory for Sonnet and Haiku, that high effort accepts fewer plausible-but-wrong patches than low in every model, and that medium matches high within the margin for Opus but not for the smaller models (Sections~\ref{sec:rq1} and~\ref{sec:rq2}).
\item A comparison of cost and agreement across nine model and effort settings that shows where the gains from effort level off in each model and which settings no other setting beats on both cost and agreement (Sections~\ref{sec:rq3} and~\ref{sec:rq5}).
\item Exploratory evidence that, in every model, the results rest on tests that may demand more than the issue asks, and that, for Opus, recall of the upstream fix goes with accepting wrong patches (Sections~\ref{sec:requires} and~\ref{sec:memory}).
\item Exploratory coding of what the 205 rejections of wrong patches and the 27 false alarms claimed, showing that most rejections named the defect the failing test checks, that effort added rejections that did so, and that most false alarms pointed at behavior the official fix shares (Section~\ref{sec:rq6}).
\end{enumerate}

\section{Background and Related Work}\label{sec:background}

\subsection{Language Models as Judges of Code}

Using a language model to grade other models' output, the LLM-as-a-judge approach, is now routine. Strong models agree with human preferences on open-ended answers about as well as humans agree with each other, with position, verbosity and self-enhancement biases \cite{zheng2023judging}. For code the evidence is mixed. Reasoning judges outperform non-reasoning ones on coding tasks \cite{jiang2026codejudgebench}; repeated runs of a judge disagree, and prompt wording can swing its verdicts \cite{zhao2026bias}; and the effect of reasoning effort on a judge differs across model families \cite{huang2025judge}. A human in the loop can make a rubric-based patch judge reliable \cite{shi2026hitl}, and verifiers that run code and verifiers that only read it have complementary strengths and work best combined \cite{jain2025r2egym}. A judge run after repair detected adversarial security patches with an average accuracy of only 55.4\% \cite{trad2026adversarial}. Our study differs in varying effort within each of three models of one family, giving the reviewer tools to run tests, and using ordinary agent patches that happen to be wrong rather than patches built to deceive.

\subsection{Weak Tests and Benchmark Labels}

A benchmark's label is only as good as its tests. On SWE-bench, patch validation lets through patches that fail the developers' own tests \cite{wang2026solved}, and tests can encode one fix's details rather than the issue's requirement \cite{ahmed2026overfitting}. SWE-ABS \cite{yu2026sweabs} generates stronger tests from the official fix, partly with a language model; its authors inspected all 500 Verified instances by hand and repaired overfit tests in 53 of them. Because the strengthened tests derive from the official fix, they are not independent truth: a test may require behavior the issue never asks for. Checking only the cases where a reviewer disagrees with a label, discrepant analysis, would overstate accuracy, because labels could then move only in the reviewer's favor \cite{hadgu1999discrepant}. We therefore checked every label, not only the discrepant ones, against written rules fixed before rescoring, and measured the strictness threat directly (RQ4).

\section{Method}\label{sec:method}

\subsection{Design and Preregistration}\label{sec:design}

The study is a within-patch experiment. For each model, one factor, effort, takes three levels: low, medium and high (the values of the CLI's \path{--effort} flag). Each of 90 patches was reviewed once at each level by each model, so each patch is its own control and all comparisons within a model are paired. For Opus, the effort order was randomized within each patch from the published seed 20261102, and the three reviews of a patch were planned to run back to back so that model drift would fall evenly across efforts. The design identifies the total effect of the effort setting on the verdict, not the route by which effort acts.

For Opus, the plan, hypotheses, margins, tests and analyses were preregistered on the Open Science Framework (OSF) on November 2, 2026, at 14:46 UTC, before the first main-run review (\url{https://osf.io/r8q2m/}). The filing attaches the study plan, the patch list, the excluded patches and the review prompt; the prompt and the patch list did not change after the filing.

The Sonnet and Haiku studies repeat the Opus design on the same 90 patches with only the model changed. Their study plan was never filed with a registry, and its hypotheses were replaced after data collection by the Opus hypotheses (Table~\ref{tab:deviations}, rows 8 and 9), so every Sonnet and Haiku result is exploratory. The supplement (\url{https://github.com/trustmybot/research-artifacts/blob/cr-effort-claude-5.5-v1.2/papers/cr-effort-claude-5.5/supplement.md}) gives the full procedure, every test and every deviation.

\subsection{Patches and Labels}\label{sec:labels}

Tasks came from SWE-bench Verified, restricted to Django, a Python web framework. Patches came from public agent submissions \cite{swebench2026experiments}, and labels from the Verified and SWE-ABS tests in the SWE-ABS verdict file \cite{openagenteval2026sweabs}:

\begin{itemize}
\item \textbf{Wrong:} passes the Verified tests and fails the SWE-ABS tests. The Verified grader would count it as solved.
\item \textbf{Correct:} passes both, on an instance that SWE-ABS strengthened.
\end{itemize}

After excluding instances used in harness and pilot checks and agents whose patch text is not public, 83 instances remained in the wrong frame and 70 in the correct frame. A seeded draw chose one patch per instance, with no agent supplying more than 10\% of a stratum: 60 wrong and 30 correct patches by these original labels, covering 68 distinct issues. Before any review, each patch ran both test suites three times locally, and all 90 final patches reproduced their original label in all three runs. The check replaced four patches from the reserve, three of them wrongly because of a parser error (Table~\ref{tab:deviations}, row 6). Supplement Sections A.1 and A.2 list every exclusion and replacement.

After data collection, each of the 90 labels was rechecked against seven written label rules, agreed before any rescoring, that ask whether the failing test checks what the issue requires and whether the patch behaves like the official fix (supplement Section A.5). Language-model agents applied the rules, and each of the three authors checked 30 of the 90 patches; both the agents and the co-authors proposed changes, and the first author ruled on each one. Eleven labels changed. Seven wrong patches became correct because their failing test checks an implementation detail of the official fix that the issue does not specify, such as a helper name or message text, and one because the test itself is broken. Three correct patches became wrong because a concrete input exists that the official fix handles and the patch does not, though no test sends it. The corrected labels, 55 wrong and 35 correct, are primary for every model. The preregistration fixed the original labels, so the correction is a deviation (Table~\ref{tab:deviations}, row 7), and Section~\ref{sec:robust} reports the original-label results for every model.

\subsection{The Reviewer}\label{sec:reviewer}

The reviewer was Claude Code 2.1.311 in non-interactive mode with model \code{claude-\allowbreak opus-5-5}, \code{claude-\allowbreak sonnet-5-5} or \code{claude-\allowbreak haiku-5-5}, in a clean configuration with no plugins or saved settings. Only the effort flag and the model differed between settings.

\begin{itemize}
\item \textbf{No internet.} The reviewer had no internet access except the Claude API. Its container sat on an internal Docker network whose only outbound route was an allowlist proxy to the model API, so it could not search the web, open the GitHub issue or download the upstream fix. Whatever it knew about a fix came from the model's training, which Section~\ref{sec:memory} measures for Opus.
\item \textbf{Input.} The verbatim SWE-bench issue text, pinned by hash, in a brief encoded in Token-Oriented Object Notation (TOON) \cite{toon}; the benchmark's hints were withheld.
\item \textbf{Code and tools.} The Django repository at the issue's base commit with the candidate patch as one anonymized commit on top. The reviewer could read files, run shell commands and write files, so it could write and run its own tests.
\item \textbf{Withheld.} The scoring tests, the official fix, the SWE-ABS tests, any label and the source agent's messages, enforced by preflight checks on the base commit before each model call.
\item \textbf{Prompt.} One generic review prompt drawn from public review guides. It fails a patch only for an unmet requirement or a functional defect that matters; style, security, performance and complexity remarks are feedback, not grounds to fail.
\item \textbf{Verdict.} A strictly decoded block with \code{pass} (accept) or \code{fail} (reject), feedback and findings. A malformed block counts as no verdict.
\end{itemize}

Supplement Section A.3 gives the full configuration.

\subsection{Procedure and Cost}\label{sec:procedure}

The Opus reviews ran on November 2 and 3, 2026, in fresh per-instance SWE-bench containers, pinned by image digest, on Google Cloud virtual machines. A \$4 budget cap and a 60-minute timeout per review guarded against runaway reviews; no review of any model reached either. The model, CLI version, environment and per-patch prompt were identical in all 270 Opus reviews; only the machine type and harness version changed after the first 12 patches (supplement Table S3, row 1). After the main run, 20 patches (11 wrong and 9 correct by the corrected labels) were reviewed again by Opus at high effort to measure run-to-run noise. Supplement Section A.4 gives commits, machines and run order.

The Sonnet and Haiku reviews ran on November 5, 2026, three days after Opus, at a later harness version that added the model flag, with the same CLI, environment file, container images, per-patch prompt and cap as the Opus run. For each patch, its six reviews (two models at three efforts) ran back to back in a shuffled order; Opus was not interleaved with them (Table~\ref{tab:deviations}, row 11). Supplement Section G gives the details.

Costs are list-price U.S. dollars that the CLI computed from token counts \cite{anthropic2026pricing}. Model API spend for the whole experiment was about \$50.69 (Table~\ref{tab:spend}). Of this, the 270 reviews analyzed per model cost \$23.30 for Opus, \$8.77 for Sonnet and \$1.23 for Haiku; Table~\ref{tab:results} gives the cost per review. The rest went to pilots, recall tests, repeated reviews, failed or uncounted attempts and the 27 later reviews of the three patches the label check wrongly replaced. Google Cloud compute was billed separately and is left out: it was not recorded per study, and the Sonnet and Haiku runs carry no study label in the billing data, so only model API spend is reported.

\begin{table}[t]
\caption{Model API Spend by Item, in USD. The 270 reviews analyzed per model are the data of every result; the other items are checks around them.}\label{tab:spend}
\small
\setlength{\tabcolsep}{3pt}
\begin{tabular}{@{}L{0.46\columnwidth}L{0.2\columnwidth}L{0.29\columnwidth}@{}}
\toprule
Item & Opus study & Sonnet and Haiku studies \\
\midrule
270 reviews analyzed per model & 23.30 & 10.00 (Sonnet 8.77, Haiku 1.23) \\
Pilot on patches outside the sample & 5.04 & 1.06 \\
Recall tests: 20 probe issues and 74 issues before the main run, 68 issues after & 5.57 (0.54 + 2.67 + 2.36) & Not run \\
Repeated reviews at high effort (20) & 3.07 & Not run \\
Failed attempts and an uncounted re-run & 0.55 & 0.01 \\
Smoke test of the harness & 0.55 & 0.00 (free probe) \\
Earliest test reviews, before the spend ledger & 0.44 & Not run \\
Stray review of a check patch & 0.11 & Not run \\
\midrule
\textbf{Study total} & \textbf{38.63} & \textbf{11.07} \\
\bottomrule
\end{tabular}

\smallskip
\raggedright\footnotesize The 27 reviews of the three wrongly replaced patches, by all three models, cost \$0.99 more, so the experiment total is \$38.63 + \$11.07 + \$0.99 = \$50.69. The Sonnet and Haiku uncounted re-run is the run record's main-run total (\$10.01) minus the analyzed reviews (\$10.0025). Sources: supplement Table S13 (Opus), Section G.5 (Sonnet and Haiku) and Section D.8 (the 27 reviews).
\end{table}

\subsection{Measures}\label{sec:measures}

\begin{itemize}
\item \textbf{False success:} the share of wrong patches the reviewer accepts.
\item \textbf{False alarm:} the share of correct patches the reviewer rejects.
\item \textbf{Balanced agreement:} the mean of (1 $-$ false success) and (1 $-$ false alarm), also known as balanced accuracy. It weights the two errors equally, and the 55/35 split of the sample does not drive it. It is not the F1 score, which ignores correct acceptances and changes with the share of wrong patches. Youden's \textit{J}, used in Section~\ref{sec:rq5}, equals twice the balanced agreement minus one. The preregistration lists balanced agreement as descriptive.
\item \textbf{True success:} the share of accepted patches that are correct, when a share \textit{w} of incoming patches is wrong; \textit{w} = .167 is the share of passing Django patches that the released SWE-ABS verdicts reject, by our count (1,162 of 6,973) \cite{openagenteval2026sweabs}.
\item \textbf{Cost} and \textbf{wall time} per review.
\item \textbf{Behavior} (output and thinking tokens, tool calls, shell test runs), used as the manipulation check.
\item \textbf{Flip rate:} the share of repeated reviews whose verdict changed (Opus only).
\end{itemize}

\subsection{Analysis}\label{sec:analysis}

\textbf{Hypothesis tests.} H2 used the paired difference in false success, medium minus high, with a Tango score 95\% confidence interval \cite{tango1998equivalence} (Newcombe interval as sensitivity \cite{newcombe1998paired}); it holds if the upper limit is below .15. H1 used a one-sided exact McNemar test on the wrong patches \cite{mcnemar1947} (mid-\textit{p} as sensitivity \cite{fagerland2013mcnemar}). H3 used the ratio of geometric-mean cost, medium over high, with a paired bias-corrected and accelerated (BCa) bootstrap \cite{efron1987bca} and a Wilcoxon signed-rank test \cite{wilcoxon1945} on log costs; it holds if the upper limit is below .70. The three tests form one family at one-sided $\alpha$ = .025 under the Holm correction \cite{holm1979sequential}, within each model. Two further rules limit any claim that medium suffices: medium may reject at most 4 of the correct patches that high accepts (guardrail), and neither medium nor high may reject more than 6 correct patches (ceiling). Both limits were set for 30 correct patches and kept unchanged on purpose for the 35 under the corrected labels, so that no threshold moved after the data were in. For Opus all of this was preregistered; the power analysis assumed 10\% to 25\% discordant patches, and a blinded look after 30 wrong patches could extend the sample to 75 (supplement Section B.3). For Sonnet and Haiku, the same tests, family, guardrail and ceiling were run on each model's 270 reviews after the data were in. An incomplete or malformed review counts as no acceptance, as the Opus plan specifies; counting it as acceptance is the sensitivity analysis.

\textbf{Descriptive and exploratory analyses.} Single proportions carry 95\% Wilson intervals \cite{wilson1927} and coder agreement is Cohen's $\kappa$ \cite{cohen1960kappa}. The comparison of the nine settings with Opus at medium uses the difference in Youden's \textit{J} with a 95\% interval from the method of variance estimates recovery (MOVER), descriptively. Three exploratory analyses address RQ4. In the \textit{issue-requirement coding}, a Claude Opus 5.5 coder judged for each wrong patch whether the issue requires the behavior the failing SWE-ABS test checks; the coding applies to every model's reviews. In the \textit{memory coding}, Claude Opus 5.5 coders blind to effort, label and cost flagged Opus reviews in which memory of the upstream fix was decisive. The \textit{recall test} is a memory probe. The model gets only the issue text, with no repository, code or tools, and is asked to write the official fix. We score the share of the official fix's changed lines that do not already appear in the issue text and that the model reproduces. A model cannot derive those lines from the issue alone, so a high score suggests that it saw the fix in training, a form of training-data contamination; models identify the files to change for SWE-bench issues from the issue text far more often than for issues from other repositories, which points the same way \cite{liang2025illusion}. The test tells us whether the reviewer may be judging a patch against a fix it remembers rather than against the issue. The memory coding and the recall test were run for Opus only. Supplement Section B defines every analysis and lists every sensitivity check.

\textbf{Findings coding (RQ6).} After all the results above were known, we coded what each rejecting review claimed. For each wrong patch, a Claude Opus 5.5 agent drafted the behaviors that make it wrong, from the failing strengthened test and the official fix, or, for the five patches shown wrong by reading the patch against the official fix, from that reading; a Claude Sonnet 5.5 agent checked each draft. Two coders, Claude Opus 5.5 (coder A) and Claude Sonnet 5.5 (coder B), both at high effort and neither shown the reviewer's model or effort, then coded every review independently in four passes: the defect type of each finding in the 232 rejecting reviews, in the scheme of M\"antyl\"a and Lassenius \cite{mantyla2009defects}, without the label or the reference behaviors; whether each of the 205 rejections of wrong patches named a reference behavior, scored named, partial or other by ReviewBench's ``same underlying issue'' rule \cite{zhou2026reviewbench}; whether the main claim of each of the 27 false alarms was real, with reason codes for findings beyond the objective of the change \cite{lin2026coderabbit} and for demands beyond the issue \cite{jin2026overcorrection}, pooled with 27 decoy rejections of wrong patches so that the coders could not tell which reviews were false alarms; and whether the feedback of the 290 accepted reviews of wrong patches raised a concern or named a reference behavior. A Claude Opus 5.5 adjudicator at medium effort, which saw the two codes in random order without coder names, settled each disagreement that the codebook sends to adjudication. Agreement is reported before adjudication as percent agreement, Cohen's $\kappa$ \cite{cohen1960kappa} and Gwet's AC1 \cite{gwet2008ac1}, with bootstrap 95\% intervals that resample reviews; shares carry Wilson intervals \cite{wilson1927}. No person validated the reference behaviors or the codes. The \href{https://github.com/trustmybot/research-artifacts/blob/cr-effort-claude-5.5-v1.1/papers/cr-effort-claude-5.5/findings/codebook.md}{codebook}, the \href{https://github.com/trustmybot/research-artifacts/tree/cr-effort-claude-5.5-v1.1/papers/cr-effort-claude-5.5/findings/prompts}{prompts} and the full method are in supplement Section I.

\subsection{Deviations From the Plans}\label{sec:deviations}

Table~\ref{tab:deviations} lists the departures from the plans that bear on how the results read. Supplement Table S3 lists all 19 in full, including four with no effect on any result (rows 1, 3, 5 and 12: a change of machine type partway through the Opus run, an unrecorded pilot decision, audit groups assigned after the run, and a sample-size check done late) and three administrative ones (rows 9, 10 and 18). None changed the Opus prompt, hypotheses or frozen patch list, though the label-check bug (row 6) affected how the list was built. RQ6 and its coding plan were added after the results were known, so RQ6 is exploratory throughout; supplement Table S26 lists the 14 departures from that plan.

\begin{table*}[t]
\caption{Deviations From the Plans. Rows 1 to 7 depart from the preregistered Opus plan and rows 8 to 12 from the unfiled Sonnet and Haiku plan.}\label{tab:deviations}
\footnotesize
\setlength{\tabcolsep}{4pt}
\begin{tabular}{@{}r L{0.26\textwidth} L{0.33\textwidth} L{0.32\textwidth}@{}}
\toprule
\# & What the plan said & What happened & Why it matters \\
\midrule
1 & The three Opus reviews of each patch run one right after another, so that any change in the model over time affects all three efforts equally. & 32 reviews failed because the account hit its API spending limit. They were re-run later in separate batches, apart from the other reviews of the same patch. & For those 32 reviews, drift over time is not balanced across efforts. The 3 failed attempts that had already cost money (\$0.55) are in the spend. \\
2 & The recall test, a memory probe in which the model writes the official fix from the issue text alone with no code or tools, runs once before the main run and is reported. & It ran before the main run as planned, but the run record did not report it and the plan's spend total left out its \$2.67. A second run after the main run was the one reported at the time. & Neither run can affect a verdict, because the reviewer never saw them. The paper uses the first run. The two runs put 9 of 67 issues in different recall groups, so groupings by recall are noisy. \\
3 & Before any review, record for each wrong patch whether the issue actually requires the behavior its failing strengthened test checks. & This was coded only after the Opus run, by one Claude Opus 5.5 coder that did not see per-patch verdicts but knew the overall results. & The split by this coding (Section~\ref{sec:requires}) is exploratory; a person should redo the coding. \\
4 & A mediation analysis \cite{imai2010mediation}: a test of whether effort lowers false success by making the reviewer check more (more thinking, tool calls and test runs). & Not done; no reason was recorded at the time. Behavior is reported only as the manipulation check (Section~\ref{sec:behavior}). In hindsight the analysis would have added little: it needs sequential ignorability \cite{imai2010mediation}, which patch difficulty likely breaks, since a harder patch draws both more checking and a different verdict; and the 14 wrong patches whose Opus verdict differed between low and high are too few to estimate an indirect effect precisely. & The paper reports that more effort came with more checking, but does not claim that the checking caused the better verdicts. \\
5 & Blind coders flag Opus reviews in which memory of the upstream fix drove the verdict. & The coders were Claude Opus 5.5, the same model under test, and no person checked their flags. & The flags are a model's judgment, and a coder of the same model may share the reviewer's blind spots about what it remembers. \\
6 & A drawn patch is replaced only if its label fails to reproduce when both test suites run locally. & A bug in our label check read only Django's per-test status lines and missed failures that Django reports only in its failure summary at the end of the log, such as a failing subtest. Three wrong patches whose label did reproduce were therefore replaced by reserve patches. & The excluded patches were not a random loss: they were one kind of wrong patch, those whose failure shows only in Django's failure summary, so the sample lacks that kind. After the main run all three models reviewed the three patches at every effort (27 reviews, supplement Section D.8); swapping them back in changes no hypothesis decision for any model (Section~\ref{sec:robust}). \\
7 & Labels are fixed before the run: wrong means the patch passes the original tests and fails the strengthened ones; correct means it passes both. & After data collection, all 90 labels were rechecked against seven written rules. Eleven changed: 8 wrong patches became correct because the failing test checks a detail the issue does not ask for, or is broken, and 3 correct patches became wrong because an input exists that the official fix handles and the patch does not. & Every rate moves, but no hypothesis decision does, for any model (Section~\ref{sec:robust}). Correcting labels after seeing the verdicts can bias them toward the reviewer; we checked every label, fixed the rules before rescoring, and labels moved both ways. \\
8 & The Sonnet and Haiku studies are preregistered before data collection. & Their written plan was approved before any paid run but never filed with a public registry. & A plan without a registry's time stamp is not a preregistration, so every Sonnet and Haiku test is exploratory. \\
9 & The Sonnet and Haiku plan tests whether Sonnet at medium, then Haiku at medium, is about as good as Opus at medium (within 15 points on Youden's \textit{J}), plus an effort trend and a coded explanation of differences. & After the data were in, these hypotheses were replaced by the three Opus hypotheses on each model. The comparison with Opus at medium is reported descriptively, and the coded explanation was not done. & No confirmatory claim compares models. Choosing analyses after seeing the data allows selective reporting, so both the plan's comparison and the Opus hypotheses are reported in full. \\
10 & A review whose verdict block is malformed is re-run once. & No review was re-run. As in the Opus plan, a malformed verdict counts as no acceptance: not a false success on a wrong patch, but a false alarm on a correct one. Counting it as acceptance is the sensitivity analysis. & 15 malformed reviews (Sonnet 10, Haiku 5) stay in the data. No hypothesis decision changes between the two ways of counting them (Section~\ref{sec:robust}). \\
11 & The Sonnet and Haiku reviews run with the same harness version as Opus. & They ran three days later, with a newer harness version, from an unmerged branch, that added the model option and fixed the proxy address. Opus was not re-run beside them. & Any change in the model service between November 2 and 5 is mixed up with the difference between models. The CLI, container images, environment, prompt and cap were the same. \\
12 & The Opus checks carry over to Sonnet and Haiku. & No sample-size check, repeated reviews, memory coding or recall test was run for Sonnet or Haiku. & Their run-to-run noise and the role of memory are unmeasured, so the recall split (Figure~\ref{fig:recall-split}) and the memory results are Opus only. \\
\bottomrule
\end{tabular}
\end{table*}

\section{Results}\label{sec:results}

\subsection{Effort Changed Reviewer Behavior in Every Model}\label{sec:behavior}

Effort changed behavior in all three models, so the manipulation check passes. Paired by patch, Opus at high produced a median of 1,710 more output tokens and 786 more thinking tokens than at low, and 2 more tool calls. Between medium and high, high produced more output tokens on 91\% of patches for Opus, 98\% for Sonnet and 97\% for Haiku, with paired median increases of 693, 829.5 and 3,445.5 tokens. Haiku changed most: at high it took a median of 8.5 turns and 7.5 tool calls against 5 and 4 at medium. Opus's median number of tool calls and shell test runs was the same at medium and high (Table~\ref{tab:results}; supplement Tables S4 and S18). No review hit the budget cap or timed out. Opus returned a usable verdict in all 270 reviews; Sonnet returned a malformed verdict in 10 and Haiku in 5.

\begin{figure*}[t]
\centering
\includegraphics[width=\textwidth]{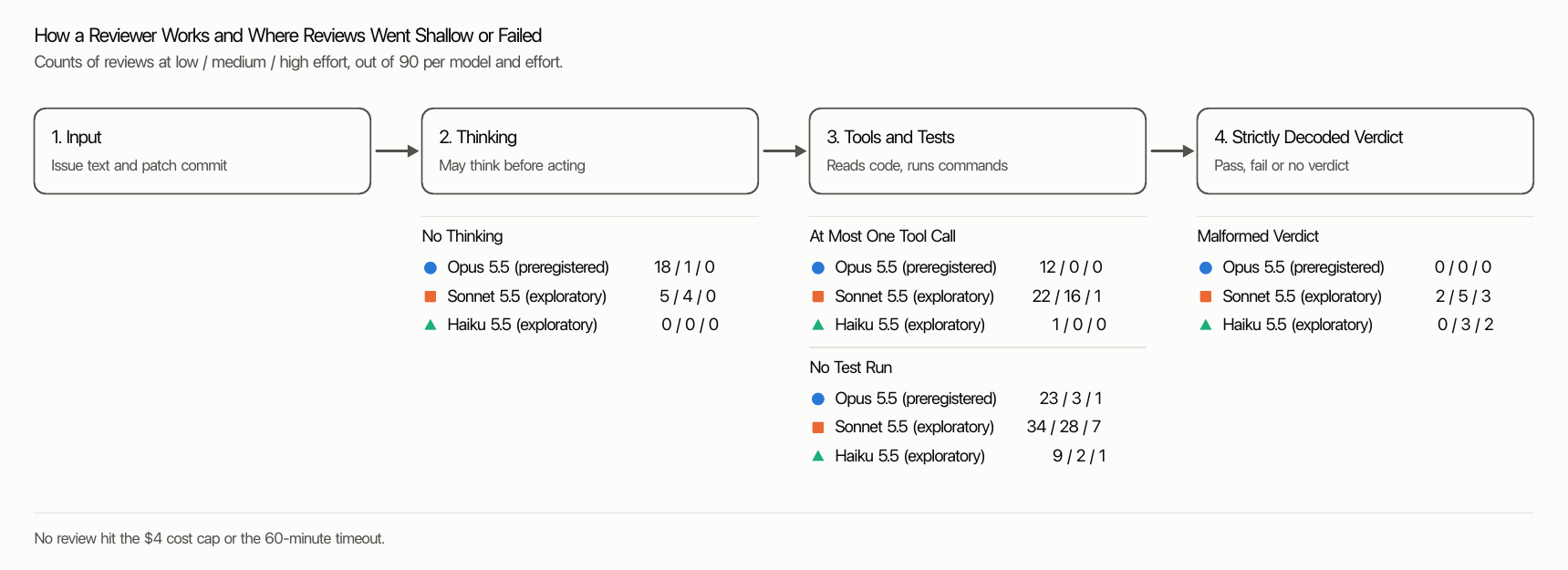}
\Description{Flowchart of the four review steps, input, thinking, tools and tests, and strictly decoded verdict, with per-model counts at low, medium and high effort of reviews with no thinking, at most one tool call, no test run and a malformed verdict.}
\caption{How a Reviewer Works and Where Reviews Went Shallow or Failed. The reviewer reads the issue and the patch, may think, reads code and runs commands and tests, and writes a verdict decoded strictly into pass, fail or no verdict. Counts are reviews at low / medium / high effort, out of 90 per setting. Opus 5.5 is preregistered; Sonnet 5.5 and Haiku 5.5 are exploratory. Shallow reviews cluster at low effort for Opus and at low and medium for Sonnet; Haiku always thought (Figure~\ref{fig:thinking}), and only the smaller models returned malformed verdicts. Source: \href{https://github.com/trustmybot/research-artifacts/blob/cr-effort-claude-5.5-v1.0/papers/cr-effort-claude-5.5/artifact/outputs/shallow-pass.json}{\code{shallow-pass.json}} and \href{https://github.com/trustmybot/research-artifacts/blob/cr-effort-claude-5.5-v1.2/papers/cr-effort-claude-5.5/artifact/outputs/models.json}{\code{models.json}} (\code{by\_model.<model>.malformed}).}\label{fig:flowchart}
\end{figure*}

\begin{figure*}[t]
\centering
\includegraphics[width=\textwidth]{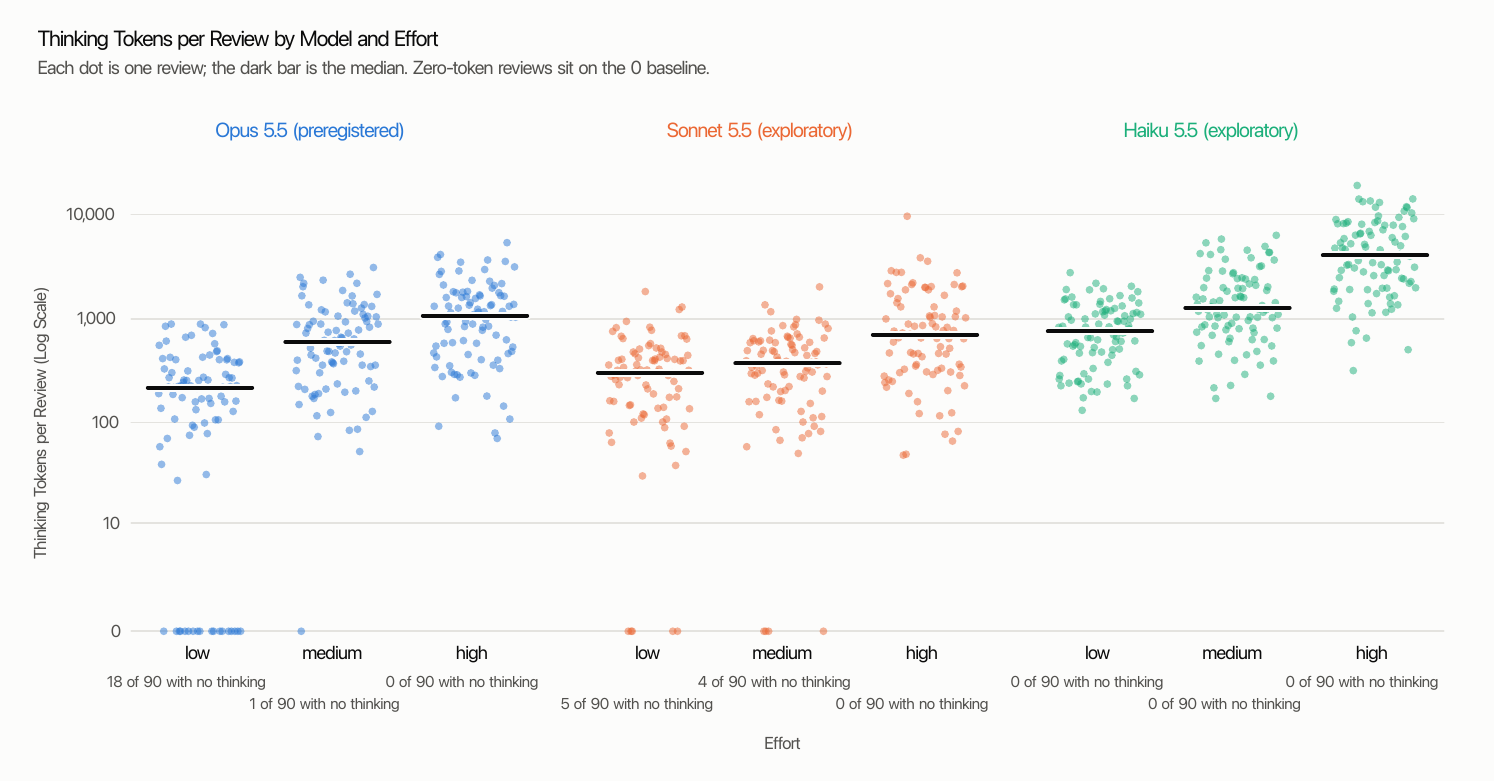}
\Description{Dot plot of thinking tokens per review on a log scale for three models at low, medium and high effort, with medians and the count of reviews with no thinking in each setting.}
\caption{Thinking Tokens per Review by Model and Effort. Each dot is one review, 90 per setting, on a log scale; the bar marks the median, and reviews with no thinking sit on the zero baseline. Opus 5.5 is preregistered; Sonnet 5.5 and Haiku 5.5 are exploratory. Haiku used at least 131 thinking tokens in every review and had the highest median at every effort, so its zero count in Figure~\ref{fig:flowchart} means it always thought, not that it reviewed more carefully. Source: \href{https://github.com/trustmybot/research-artifacts/blob/cr-effort-claude-5.5-v1.0/papers/cr-effort-claude-5.5/haiku-vs-opus/artifact/data/reviews.csv}{\code{reviews.csv}} (\code{thinking\_tokens}).}\label{fig:thinking}
\end{figure*}

Some passes at low effort came with almost no checking (exploratory count, supplement Section H; Figure~\ref{fig:flowchart}). Every review that used no thinking tokens returned a pass, and 12 of the 33 correct patches Opus accepted at low, with 6 of the 37 wrong ones, were accepted without thinking. Sonnet skipped thinking in 5 reviews at low and 4 at medium. Haiku never skipped thinking (Figure~\ref{fig:thinking}), but Haiku and Sonnet were the only models to return malformed verdicts. The vendor notes that at lower effort the model can skip thinking on problems it judges simple \cite{anthropic2026effort}. We have not read the reasons these reviews gave.

\begin{table*}[t]
\caption{Results by Model and Effort. Opus 5.5 is the preregistered study; Sonnet 5.5 and Haiku 5.5 are exploratory. Rates count a malformed review as no acceptance. Costs are list-price USD per review; wall time (minutes) and behavior values are medians per review. More effort cost more and caught more wrong patches in every model; for Opus the gain came mostly between low and medium.}\label{tab:results}
\footnotesize
\setlength{\tabcolsep}{3pt}
\begin{tabular}{@{}l R{1.25cm} R{1.25cm} R{1.15cm} R{0.95cm} R{1.2cm} R{0.8cm} R{1.0cm} R{1.0cm} R{1.3cm} R{0.75cm} R{0.85cm} R{1.15cm}@{}}
\toprule
Effort & False success (55 wrong) & False alarm (35 correct) & Balanced agreement & Mean cost & Geometric-mean cost & Wall time & Output tokens & Thinking tokens & Reviews with any thinking & Tool calls & Shell test runs & Malformed verdicts \\
\midrule
\multicolumn{13}{@{}l}{\textit{Opus 5.5 (preregistered)}} \\
Low & 37 (.67) & 2 (.06) & .635 & 0.0441 & 0.0397 & 0.37 & 1,106 & 216 & 72 & 3 & 1 & 0 \\
Medium & 26 (.47) & 2 (.06) & .735 & 0.0946 & 0.0805 & 0.65 & 2,243.5 & 598 & 89 & 4 & 2 & 0 \\
High & 25 (.45) & 2 (.06) & .744 & 0.1202 & 0.1009 & 0.82 & 2,948.5 & 1,068 & 90 & 4 & 2 & 0 \\
\addlinespace
\multicolumn{13}{@{}l}{\textit{Sonnet 5.5 (exploratory)}} \\
Low & 36 (.65) & 2 (.06) & .644 & 0.0232 & 0.0210 & 0.28 & 1,136.5 & 303 & 85 & 2.5 & 1 & 2 \\
Medium & 33 (.60) & 3 (.09) & .657 & 0.0272 & 0.0241 & 0.32 & 1,216.5 & 372.5 & 86 & 2 & 1 & 5 \\
High & 27 (.49) & 3 (.09) & .712 & 0.0470 & 0.0392 & 0.50 & 2,063 & 699 & 90 & 4 & 1 & 3 \\
\addlinespace
\multicolumn{13}{@{}l}{\textit{Haiku 5.5 (exploratory)}} \\
Low & 43 (.78) & 2 (.06) & .581 & 0.0023 & 0.0021 & 0.32 & 2,198.5 & 756 & 90 & 3 & 1 & 0 \\
Medium & 35 (.64) & 7 (.20) & .582 & 0.0037 & 0.0031 & 0.46 & 3,226 & 1,268.5 & 90 & 4 & 1 & 3 \\
High & 28 (.51) & 4 (.11) & .688 & 0.0078 & 0.0063 & 0.99 & 7,055 & 4,063 & 90 & 7.5 & 1 & 2 \\
\bottomrule
\end{tabular}

\smallskip
\raggedright\footnotesize \textit{n} = 90 reviews per row. A review without thinking produced no thinking block and counts as 0 thinking tokens; the zero is measured, not missing. Wilson intervals for the rates are in Figure~\ref{fig:error-rates} and supplement Table S15. Sources: \url{https://github.com/trustmybot/research-artifacts/blob/cr-effort-claude-5.5-v1.2/papers/cr-effort-claude-5.5/artifact/outputs/models.json} (\code{by\_model.<model>.rates}, \code{cost}, \code{manipulation}, \code{malformed}; its Opus rows reproduce \url{https://github.com/trustmybot/research-artifacts/blob/cr-effort-claude-5.5-v1.2/papers/cr-effort-claude-5.5/artifact/outputs/results.json}) and \url{https://github.com/trustmybot/research-artifacts/blob/cr-effort-claude-5.5-v1.0/papers/cr-effort-claude-5.5/artifact/outputs/shallow-pass.json} (\code{by\_model.<model>.<effort>.no\_thinking}). Full tables: supplement Sections D and G.
\end{table*}

\subsection{RQ1: High Effort Caught More Wrong Patches Than Low in Every Model}\label{sec:rq1}

\begin{figure*}[t]
\centering
\includegraphics[width=\textwidth]{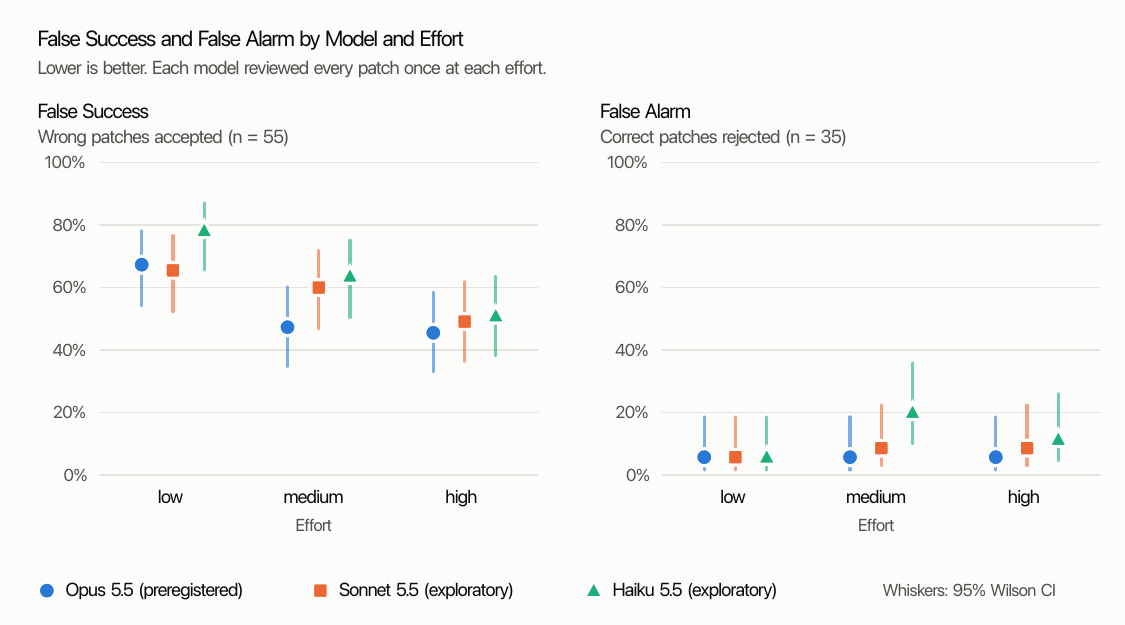}
\Description{Dot plot of false success and false alarm with 95\% Wilson intervals for three models at low, medium and high effort.}
\caption{False Success and False Alarm by Model and Effort. Left: share of the 55 wrong patches accepted; right: share of the 35 correct patches rejected. Opus 5.5 is preregistered; Sonnet 5.5 and Haiku 5.5 are exploratory. Whiskers are 95\% Wilson intervals; lower is better. Effort lowered false success in every model, for Opus mostly from low to medium, and false alarm stayed far below false success. Source: \href{https://github.com/trustmybot/research-artifacts/blob/cr-effort-claude-5.5-v1.2/papers/cr-effort-claude-5.5/artifact/outputs/models.json}{\code{models.json}} (\code{by\_model.<model>.rates}).}\label{fig:error-rates}
\end{figure*}

In every model, high effort accepted fewer wrong patches than low: false success fell from .67 to .45 for Opus, from .65 to .49 for Sonnet and from .78 to .51 for Haiku (Table~\ref{tab:results}, Figure~\ref{fig:error-rates}). For Opus, H1 is supported in the preregistered test: high caught 13 wrong patches that low accepted, and low caught 1 that high accepted (exact McNemar \textit{p} $<$ .001; Table~\ref{tab:hypotheses}). In the exploratory tests, H1 is supported for Haiku (16 against 1, \textit{p} $<$ .001). For Sonnet (12 against 3), the exact test (\textit{p} = .018) misses its Holm threshold of .0125, though the mid-\textit{p} test (\textit{p} = .011) passes it.

False alarm moved little with effort. Opus rejected 2 of the 35 correct patches at every effort and Sonnet 2 or 3. Haiku rejected 2 at low, 7 at medium and 4 at high, the only setting near the ceiling of 6.

\subsection{RQ2: Medium Matched High for Opus Only}\label{sec:rq2}

\begin{figure*}[t]
\centering
\includegraphics[width=\textwidth]{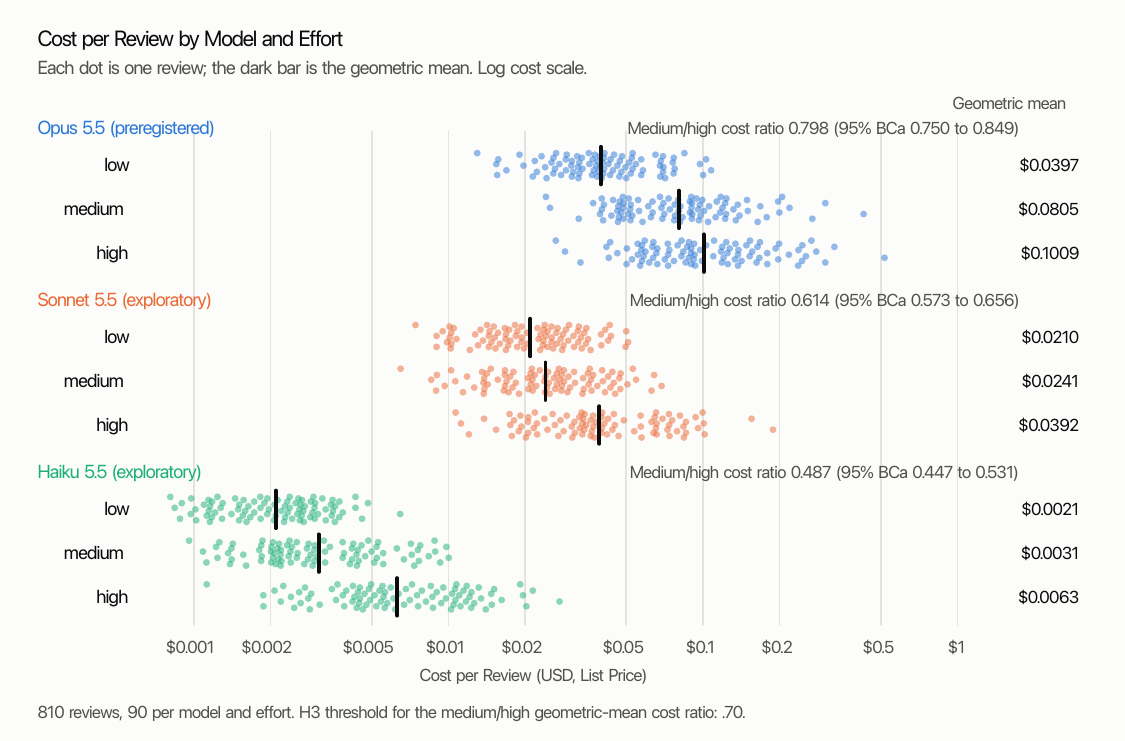}
\Description{Strip plot of 810 review costs on a log scale for three models at low, medium and high effort, with geometric means marked for each setting.}
\caption{Cost per Review by Model and Effort. Each dot is one review, 90 per setting, on a log scale; the bar marks the geometric mean, and each panel gives the medium-to-high cost ratio with its 95\% BCa interval. Opus 5.5 is preregistered; Sonnet 5.5 and Haiku 5.5 are exploratory. Only Opus stayed above the preregistered threshold of .70: medium saved more, relative to high, in the smaller models. Source: \href{https://github.com/trustmybot/research-artifacts/blob/cr-effort-claude-5.5-v1.0/papers/cr-effort-claude-5.5/haiku-vs-opus/artifact/data/reviews.csv}{\code{reviews.csv}} (\code{cost\_usd}) and \href{https://github.com/trustmybot/research-artifacts/blob/cr-effort-claude-5.5-v1.2/papers/cr-effort-claude-5.5/artifact/outputs/models.json}{\code{models.json}} (\code{by\_model.<model>.hypotheses.H3}).}\label{fig:cost}
\end{figure*}

For Opus, medium was non-inferior to high, but its cost saving fell short of the preregistered threshold. H2 is supported: medium and high disagreed on only 3 of the 55 wrong patches, 2 accepted by medium alone and 1 by high alone, so the difference in false success is .018 (95\% CI [$-$.065, .108]), inside the 15-point margin (Table~\ref{tab:hypotheses}). Neither limit on the medium conclusion was triggered: medium rejected no correct patch that high accepted (guardrail 4), and each effort had 2 false alarms (ceiling 6). H3 is not supported: medium cost 20\% less than high in geometric mean (ratio .798, 95\% CI [.750, .849]; Figure~\ref{fig:cost}), but the whole interval lies above the .70 threshold for a material saving. Under the Holm correction, H1 and H2 are supported and H3 is not, with the mid-\textit{p} McNemar test as well.

For Sonnet and Haiku the pattern reverses. Medium cost 39\% less than high for Sonnet (ratio .614, 95\% CI [.573, .656]) and 51\% less for Haiku (.487 [.447, .531]), so H3 is supported for both. But medium and high disagreed on many more wrong patches, 10 for Sonnet and 15 for Haiku, mostly patches that medium accepted and high rejected (8 against 2 and 11 against 4). The H2 intervals reach .234 and .269, above the 15-point margin, so medium was not shown non-inferior to high in either model. Sonnet stayed within both limits: medium rejected 2 correct patches that high accepted, and the most false alarms in one setting was 3. Haiku breached both: medium rejected 5 correct patches that high accepted, against a guardrail of 4, and gave 7 false alarms, above the ceiling of 6.

\begin{table*}[t]
\caption{Hypothesis Tests by Model. Tests on the corrected labels; the Opus tests are preregistered and confirmatory, and the Sonnet and Haiku tests exploratory. H1 and H2 use the 55 wrong patches; H3 uses all 90. \textit{b} and \textit{c} count discordant wrong patches (H1: low accepts and high rejects, and the reverse; H2: medium accepts and high rejects, and the reverse). The Holm correction runs within each model at one-sided $\alpha$ = .025, and the threshold depends on the order of the \textit{p} values in that model. The guardrail and the ceiling limit any claim that medium suffices.}\label{tab:hypotheses}
\footnotesize
\begin{tabular}{@{}L{4.6cm}L{3.9cm}lL{2.3cm}lL{1.8cm}@{}}
\toprule
Hypothesis or limit & Estimate & 95\% CI & One-sided \textit{p} & Holm threshold & Supported \\
\midrule
\multicolumn{6}{@{}l}{\textit{Opus 5.5 (preregistered)}} \\
H1: high false success $<$ low & Difference $-$.218 (\textit{b} = 13, \textit{c} = 1) & [$-$.350, $-$.101] & .0009 (exact McNemar) & .0083 & Yes \\
H2: medium false success $\leq$ high + .15 & Difference .018 (\textit{b} = 2, \textit{c} = 1) & [$-$.065, .108] & .0052 (Tango) & .0125 & Yes \\
H3: cost ratio medium/high $<$ .70 & Ratio .798 & [.750, .849] & $>$ .999 (Wilcoxon) & .025 & No \\
Guardrail: medium rejects $\leq$ 4 correct patches that high accepts & 0 & & & & Held \\
Ceiling: $\leq$ 6 false alarms at medium and high & 2 and 2 & & & & Held \\
\addlinespace
\multicolumn{6}{@{}l}{\textit{Sonnet 5.5 (exploratory)}} \\
H1: high false success $<$ low & Difference $-$.164 (\textit{b} = 12, \textit{c} = 3) & [$-$.303, $-$.029] & .0176 (exact McNemar) & .0125 & No (yes with mid-\textit{p}) \\
H2: medium false success $\leq$ high + .15 & Difference .109 (\textit{b} = 8, \textit{c} = 2) & [$-$.004, .234] & .2406 (Tango) & .025 & No \\
H3: cost ratio medium/high $<$ .70 & Ratio .614 & [.573, .656] & .0002 (Wilcoxon) & .0083 & Yes \\
Guardrail & 2 & & & & Held \\
Ceiling & 3 and 3 & & & & Held \\
\addlinespace
\multicolumn{6}{@{}l}{\textit{Haiku 5.5 (exploratory)}} \\
H1: high false success $<$ low & Difference $-$.273 (\textit{b} = 16, \textit{c} = 1) & [$-$.408, $-$.147] & .0001 (exact McNemar) & .0125 & Yes \\
H2: medium false success $\leq$ high + .15 & Difference .127 (\textit{b} = 11, \textit{c} = 4) & [$-$.012, .269] & .3704 (Tango) & .025 & No \\
H3: cost ratio medium/high $<$ .70 & Ratio .487 & [.447, .531] & $<$ .0001 (Wilcoxon) & .0083 & Yes \\
Guardrail & 5 & & & & Breached \\
Ceiling & 7 and 4 & & & & Breached \\
\bottomrule
\end{tabular}

\smallskip
\raggedright\footnotesize Sources: \url{https://github.com/trustmybot/research-artifacts/blob/cr-effort-claude-5.5-v1.2/papers/cr-effort-claude-5.5/artifact/outputs/models.json} (\code{by\_model.<model>.hypotheses}, \code{holm}, \code{holm\_with\_mid\_p}, \code{guardrail}, \code{ceiling}) and, for Opus, \url{https://github.com/trustmybot/research-artifacts/blob/cr-effort-claude-5.5-v1.2/papers/cr-effort-claude-5.5/artifact/outputs/results.json} (\code{hypotheses}, \code{holm}). Test statistics and sensitivity intervals: supplement Tables S6 and S16.
\end{table*}

\subsection{RQ3 (Exploratory): Gains From Effort Leveled Off by Medium for Opus Only}\label{sec:rq3}

\begin{figure*}[t]
\centering
\includegraphics[width=\textwidth]{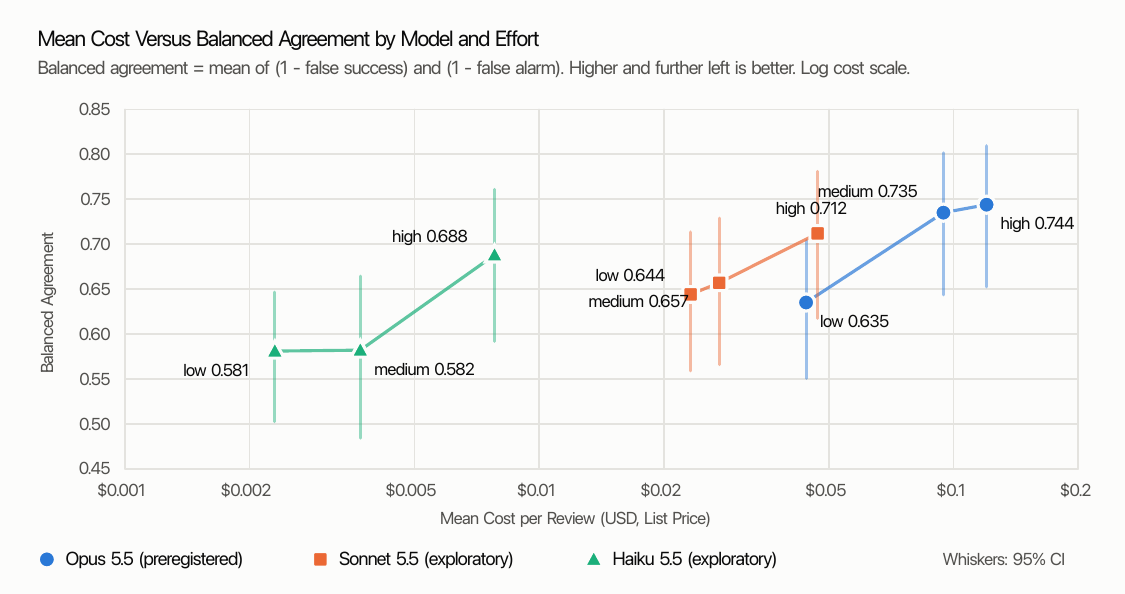}
\Description{Scatter of nine model and effort settings plotting mean cost per review on a log scale against balanced agreement.}
\caption{Mean Cost Versus Balanced Agreement by Model and Effort. One point per setting, with cost on a log scale; whiskers are 95\% intervals from Youden's \textit{J}, and higher and further left is better. Opus 5.5 is preregistered; Sonnet 5.5 and Haiku 5.5 are exploratory. Opus at medium and high agree most; Haiku at high agrees more than Opus at low and any Sonnet setting below high, at a fraction of their cost. Source: \href{https://github.com/trustmybot/research-artifacts/blob/cr-effort-claude-5.5-v1.2/papers/cr-effort-claude-5.5/artifact/outputs/models.json}{\code{models.json}} (\code{by\_model.<model>.rates}, \code{cost}, \code{youden}).}\label{fig:cost-agreement}
\end{figure*}

For Opus almost all of the gain in agreement came from low to medium; for Sonnet and Haiku most of it came from medium to high (Table~\ref{tab:results}, Figure~\ref{fig:cost-agreement}). Balanced agreement rose from .635 to .735 and .744 for Opus, from .644 to .657 and .712 for Sonnet, and from .581 to .582 and .688 for Haiku.

For Opus, moving from low to medium cost \$0.051 more per review and caught 11 more wrong patches, or \$0.41 per extra wrong patch caught, with no extra false alarm. Moving from medium to high cost \$0.026 more per review, 27\% more in mean cost, and caught 1 more wrong patch, or \$2.30 per extra catch, again with no extra false alarm. For Sonnet, low to medium cost \$0.0040 more per review and caught 3 more wrong patches with 1 more false alarm, and medium to high cost \$0.0198 more and caught 6 more with no extra false alarm (\$0.121 and \$0.296 per extra catch for the two steps). For Haiku, low to medium cost \$0.0014 more per review and caught 8 more wrong patches but added 5 false alarms, and medium to high cost \$0.0041 more, caught 7 more and had 3 fewer false alarms (\$0.016 and \$0.053 per extra catch). In absolute terms, every step up in effort for Haiku cost less than a cent per review.

The reviewer improves the stream of accepted patches only modestly. If 16.7\% of incoming patches are wrong, as for Django in the released SWE-ABS verdicts \cite{openagenteval2026sweabs}, 83.3\% of patches are correct with no reviewer. Among the patches the reviewer accepts, 87.5\% are correct for Opus at low, 90.9\% at medium and 91.2\% at high; for Sonnet 87.8\%, 88.4\% and 90.3\% and for Haiku 85.7\%, 86.2\% and 89.7\% at low, medium and high. If half the patches are wrong, the Opus shares are 58.4\%, 66.6\% and 67.5\%.

\subsection{Robustness of the Hypothesis Decisions}\label{sec:robust}

In every model, the hypothesis decisions held under the original labels, with the three wrongly replaced patches swapped back in, and with malformed verdicts counted as acceptances (Table~\ref{tab:robust}). Repeated Opus reviews did not flip, and plausible label error kept the ordering of the efforts.

Under the original, preregistered labels (60 wrong and 30 correct), Opus false success was .72 at low and .55 at both medium and high, and false alarm 3, 4 and 5 of 30. H1 is supported (difference $-$.167, exact \textit{p} = .0032), H2 is supported (difference 0, 95\% CI [$-$.074, .074], \textit{p} = .001) and H3 is not. Balanced agreement was .592, .658 and .642, so under the original labels medium, not high, agreed most for Opus. Sonnet's balanced agreement under the original labels was highest at low (.625), and Haiku's at high (.633). The decisions for Sonnet (H3 only) and Haiku (H1 and H3) are the same as under the corrected labels.

The three patches that the faulty label check replaced (Table~\ref{tab:deviations}, row 6), wrong by the original labels, were reviewed afterwards by all three models at every effort, and all 27 reviews accepted the patch. Under the corrected labels one of the three is correct; swapping all three in raises Opus false success at medium from .473 to .491 and Haiku's at high from .509 to .527, leaves Sonnet's unchanged, and changes no hypothesis decision for any model. Counting the 15 malformed verdicts as acceptances also changes no decision. None of the 20 repeated Opus reviews at high effort changed its verdict (flip rate 0, 95\% Wilson CI [0, .161]), so the data bound noise only loosely, at up to about 16\%. If 20\% of the wrong labels were in fact correct patches, Opus false success on truly wrong patches would be .61 at low, .36 at medium and .33 at high, so the ordering holds. Supplement Sections D.8, D.10 and G give these and further checks.

\begin{table*}[t]
\caption{Robustness Checks by Model. Opus 5.5 is preregistered; Sonnet 5.5 and Haiku 5.5 are exploratory. Rates are false success at low / medium / high unless stated. ``Decisions'' lists the hypotheses supported under Holm; on the corrected labels they are H1 and H2 for Opus, H3 for Sonnet, and H1 and H3 for Haiku.}\label{tab:robust}
\small
\begin{tabular}{@{}L{4.6cm}L{3.9cm}L{3.9cm}L{3.9cm}@{}}
\toprule
Check & Opus 5.5 (preregistered) & Sonnet 5.5 (exploratory) & Haiku 5.5 (exploratory) \\
\midrule
Original labels: false success (60 wrong) & 43 / 33 / 33 (.72 / .55 / .55) & 41 / 39 / 35 (.68 / .65 / .58) & 48 / 39 / 34 (.80 / .65 / .57) \\
Original labels: false alarm (30 correct) & 3 / 4 / 5 & 2 / 4 / 6 & 2 / 6 / 5 \\
Original labels: balanced agreement & .592 / .658 / .642 & .625 / .608 / .608 & .567 / .575 / .633 \\
Original labels: decisions & H1, H2 (unchanged) & H3 (unchanged) & H1, H3 (unchanged) \\
Three replaced patches swapped back in & .673 / .491 / .473; decisions unchanged & .655 / .600 / .491; decisions unchanged & .782 / .636 / .527; decisions unchanged \\
Malformed verdicts counted as acceptance & No malformed verdicts & Decisions unchanged (H1 exact \textit{p} = .0287) & Decisions unchanged (H2 \textit{p} = .3612) \\
Mid-\textit{p} McNemar for H1 & Decisions unchanged & H1 also supported & Decisions unchanged \\
Repeated reviews at high (20 patches) & 0 of 20 flipped [0, .161] & Not run & Not run \\
Label error at 20\%: corrected false success & .605 / .355 / .332 & .582 / .521 / .385 & .742 / .595 / .415 \\
\bottomrule
\end{tabular}

\smallskip
\raggedright\footnotesize Sources: \url{https://github.com/trustmybot/research-artifacts/blob/cr-effort-claude-5.5-v1.2/papers/cr-effort-claude-5.5/artifact/outputs/models.json} (\code{original\_labels.by\_model.<model>}: \code{rates}, \code{rates\_malformed\_as\_accept}, \code{holm\_malformed\_as\_accept} and \code{holm\_with\_mid\_p}, and for Sonnet and Haiku the original-label decisions in \code{hypotheses}, \code{holm}, \code{guardrail} and \code{ceiling}), \url{https://github.com/trustmybot/research-artifacts/blob/cr-effort-claude-5.5-v1.2/papers/cr-effort-claude-5.5/artifact/outputs/models-swap.json} (\code{by\_model}, for the swap only; supplement Section D.8) and \url{https://github.com/trustmybot/research-artifacts/blob/cr-effort-claude-5.5-v1.2/papers/cr-effort-claude-5.5/artifact/outputs/results.json} (\code{original\_labels}, \code{repeats}, \code{label\_error\_sensitivity}).
\end{table*}

\subsection{RQ4 (Exploratory): Results Depend on Possibly Over-Strict Tests}\label{sec:requires}

In every model, the results rest largely on wrong patches whose failing test the issue may not require. The post hoc coder judged that the issue requires the behavior of the failing SWE-ABS test for only 6 of the 55 wrong patches, does not require it for 32 and is unclear for 14. The other 3 are the patches the label correction made wrong; no test fails them, so they were not coded. In every setting, false success was lowest where the issue requires the tested behavior and was high where it does not (Table~\ref{tab:requires}). Most of each model's gain from effort came on the not-required patches: from low to high, false success there fell by 7 patches for Opus, 6 for Sonnet and 9 for Haiku.

\begin{table}[t]
\caption{Wrong Patches Accepted by Whether the Issue Requires the Tested Behavior. Post hoc coding by one language-model coder of whether the issue requires the behavior the failing SWE-ABS test checks. Cells give the count accepted and the rate. Opus 5.5 is preregistered; Sonnet 5.5 and Haiku 5.5 are exploratory, and this split is exploratory for every model. Most false successes fall on patches whose failing test the issue does not require.}\label{tab:requires}
\footnotesize
\setlength{\tabcolsep}{2.5pt}
\begin{tabular}{@{}l R{1.05cm} R{1.1cm} R{1.15cm} R{1.15cm} R{1.2cm}@{}}
\toprule
Effort & Required (6) & Unclear (14) & Not required (32) & No failing test (3) & All wrong patches (55) \\
\midrule
\multicolumn{6}{@{}l}{\textit{Opus 5.5 (preregistered)}} \\
Low & 2 (.33) & 9 (.64) & 24 (.75) & 2 (.67) & 37 (.67) \\
Medium & 1 (.17) & 7 (.50) & 17 (.53) & 1 (.33) & 26 (.47) \\
High & 1 (.17) & 7 (.50) & 17 (.53) & 0 (0) & 25 (.45) \\
\addlinespace
\multicolumn{6}{@{}l}{\textit{Sonnet 5.5 (exploratory)}} \\
Low & 1 (.17) & 8 (.57) & 24 (.75) & 3 (1.00) & 36 (.65) \\
Medium & 1 (.17) & 8 (.57) & 22 (.69) & 2 (.67) & 33 (.60) \\
High & 1 (.17) & 8 (.57) & 18 (.56) & 0 (0) & 27 (.49) \\
\addlinespace
\multicolumn{6}{@{}l}{\textit{Haiku 5.5 (exploratory)}} \\
Low & 5 (.83) & 9 (.64) & 26 (.81) & 3 (1.00) & 43 (.78) \\
Medium & 3 (.50) & 7 (.50) & 23 (.72) & 2 (.67) & 35 (.64) \\
High & 2 (.33) & 7 (.50) & 17 (.53) & 2 (.67) & 28 (.51) \\
\bottomrule
\end{tabular}

\smallskip
\raggedright\footnotesize Sources: \url{https://github.com/trustmybot/research-artifacts/blob/cr-effort-claude-5.5-v1.2/papers/cr-effort-claude-5.5/artifact/outputs/models.json} (\code{by\_model.<model>.label\_question.by\_group}), \url{https://github.com/trustmybot/research-artifacts/blob/cr-effort-claude-5.5-v1.2/papers/cr-effort-claude-5.5/artifact/outputs/results.json} (\code{label\_question}) and \url{https://github.com/trustmybot/research-artifacts/blob/cr-effort-claude-5.5-v1.0/papers/cr-effort-claude-5.5/artifact/data/issue-requires-coding.csv}.
\end{table}

Without the 32 not-required patches, 23 wrong patches remain, and no model shows that high beats low at the .025 level or that medium is non-inferior to high. The high-minus-low difference is $-$.217 for Opus (95\% CI [$-$.437, .010], exact \textit{p} = .063, mid-\textit{p} = .035), $-$.130 for Sonnet ([$-$.361, .108], \textit{p} = .23) and $-$.261 for Haiku ([$-$.481, $-$.024], \textit{p} = .035). The upper limits of the H2 intervals, .236, .291 and .282, all exceed the margin. The cost ratios barely move (.806, .611 and .505). The smaller sample explains part of the loss, but on this reading some of the reviewers' false successes may be correct acceptances.

\subsection{RQ4 (Exploratory): Fix Recall Went With Accepting Wrong Patches for Opus}\label{sec:memory}

The memory coding and the recall test were run for Opus only (Table~\ref{tab:deviations}, row 12). Opus reviews often drew on memory of the upstream fix, and those reviews accepted wrong patches more often. The coders flagged 107 of the 270 Opus reviews as ones where memory of the fix was decisive in the thinking or appeared in the final reasons: 41 at low, 37 at medium and 29 at high (agreement on 60 double-coded reviews, $\kappa$ = .77). Flagged reviews accepted 34 of 48 wrong patches (.71) against 54 of 117 (.46) for unflagged ones, while false alarms stayed rare in both groups (supplement Section D.6). This is an association, which patch difficulty could also explain. When the preregistered sensitivity analysis drops every patch with a flagged review, 32 patches remain (25 wrong and 7 correct) and H1 and H2 point the same way as in the full sample (supplement Section D.6).

\begin{figure*}[t]
\centering
\includegraphics[width=\textwidth]{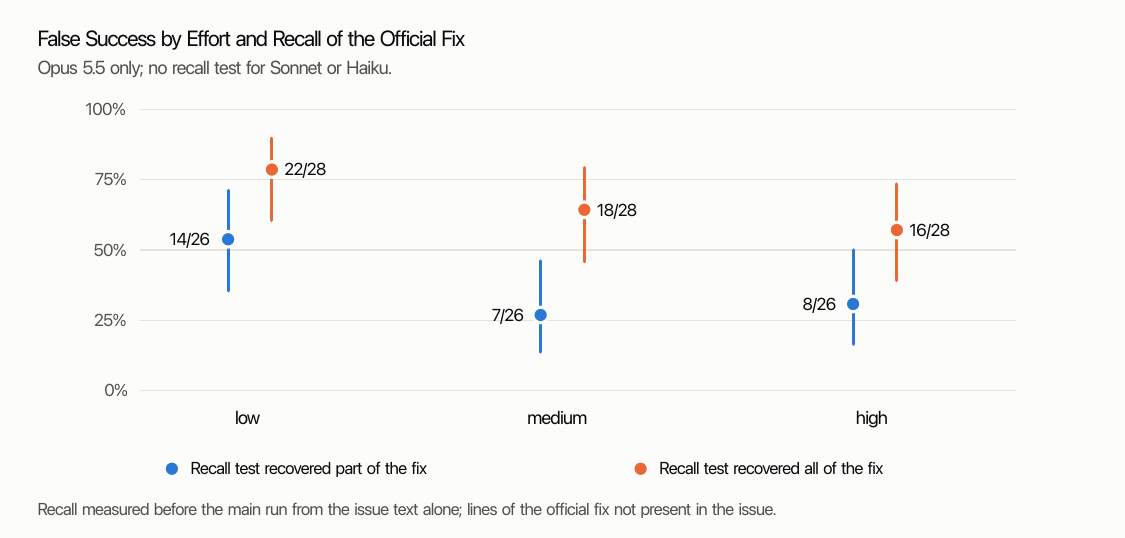}
\Description{Dot plot of false success at three efforts for issues with partial and full recall of the official fix, Opus only.}
\caption{False Success by Effort and Recall of the Official Fix, Opus 5.5 Only. Exploratory split of the wrong patches by whether the recall test recovered all (28) or only part (26) of the official fix's lines that are not in the issue text; one patch whose fix has no such lines is left out. No recall test was run for Sonnet or Haiku (Table~\ref{tab:deviations}, row 12). Whiskers are 95\% Wilson intervals. Wrong patches on fully recalled issues were accepted more often at every effort. Source: \href{https://github.com/trustmybot/research-artifacts/blob/cr-effort-claude-5.5-v1.2/papers/cr-effort-claude-5.5/artifact/outputs/results.json}{\code{results.json}} (\code{recall.split}).}\label{fig:recall-split}
\end{figure*}

The recall test confirms that Opus knows many of these fixes. From the issue text alone, before the main run, it reproduced 67.0\% of the official fixes' changed lines that do not appear in the issue, and all of them for 38 of 67 issues. A second run after the main run recovered 73.5\% and regrouped 9 of the 67 issues ($\kappa$ = .72; supplement Table S8), so recall of a given fix varies between runs. Wrong patches on fully recalled issues were accepted more often at every effort than those on partially recalled issues, and effort lowered false success in both groups (Figure~\ref{fig:recall-split}). Because 7 of the 54 grouped wrong patches would change group under the second run, the split is descriptive only.

\subsection{RQ5 (Exploratory): Opus at Medium or High Agreed Best; Haiku at High Was the Cheapest Strong Setting}\label{sec:rq5}

No setting was best on both cost and agreement, but six of the nine settings were not beaten on both by any other (Figure~\ref{fig:cost-agreement}, Table~\ref{tab:results}). In order of cost these are Haiku at low, medium and high, Sonnet at high, and Opus at medium and high. Each of the other three, Sonnet at low and at medium and Opus at low, cost more per review than Haiku at high and agreed less with the labels. Opus at high had the highest balanced agreement (.744) and Youden's \textit{J} (.488), just above Opus at medium (.735 and .470).

The study plan's reference was Opus at medium, the default effort of the strongest model. Against it, no other model's setting was shown non-inferior on the difference in \textit{J} with a 15-point margin: the lower limits run from $-$.51 to $-$.23 (supplement Table S17). Two settings came closest at much lower cost. Sonnet at high cost about half as much per review as Opus at medium (\$0.0470 against \$0.0946) and had a difference in \textit{J} of $-$.047 (95\% CI [$-$.229, .132]). Haiku at high cost about 8\% as much (\$0.0078) and had a difference in \textit{J} of $-$.094 (95\% CI [$-$.285, .094]). Both intervals are too wide to call either setting as good as Opus at medium or worse than it.

The patch-by-patch comparison shows how close the catches were. Opus at medium turned away 29 of the 55 wrong patches, Sonnet at high 28 and Haiku at high 27. Haiku at high and Opus at medium both turned away 22; Haiku at high alone turned away 5 and Opus at medium alone 7 (exact \textit{p} = .77; supplement Table S17). A separate exploratory note (\href{https://github.com/trustmybot/research-artifacts/blob/cr-effort-claude-5.5-v1.2/papers/cr-effort-claude-5.5/haiku-vs-opus/analysis.md}{\code{haiku-vs-opus/analysis.md}}) finds that which patch is under review explains far more of the catches than the model and effort setting ($\eta^2$ = .56 against .05). But Haiku at high breached the guardrail and the ceiling at medium, gave 4 false alarms against Opus's 2, and returned 2 malformed verdicts, so a gate built on it needs a fallback for unusable reviews. Per dollar, Haiku at high was the strongest setting in this experiment; per wrong patch caught, Opus at medium or high remained the safest choice.

\subsection{RQ6 (Exploratory): Effort Added Rejections That Named the Tested Defect}\label{sec:rq6}

\begin{figure*}[t]
\centering
\includegraphics[width=\textwidth]{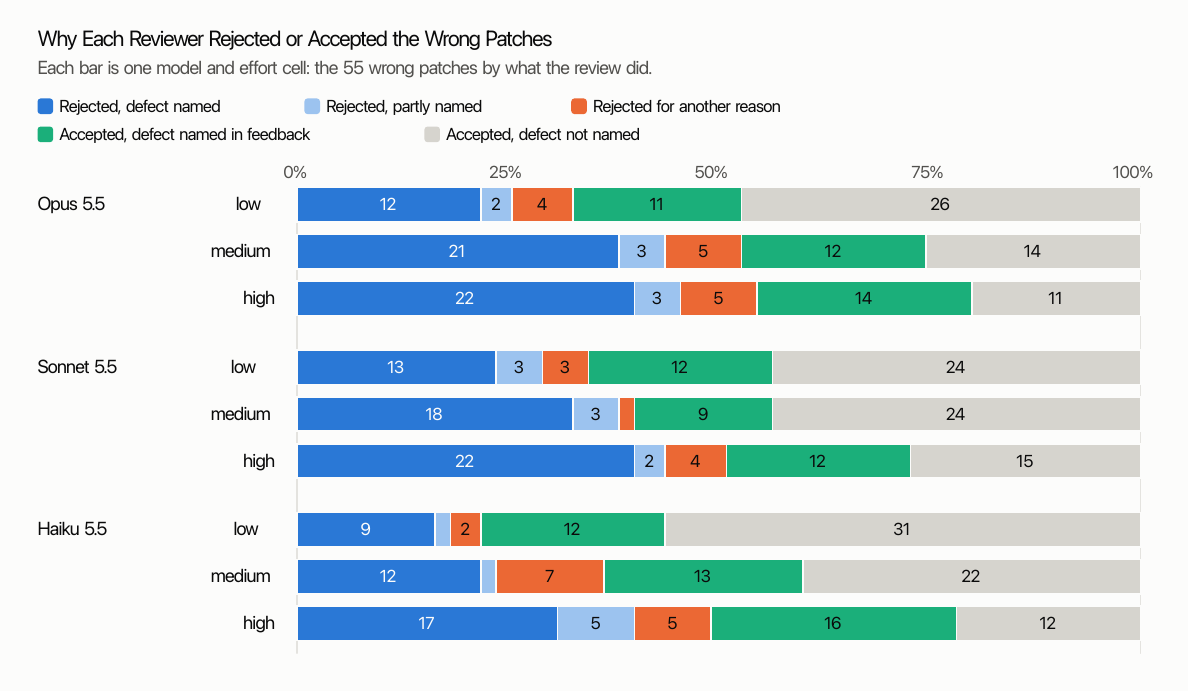}
\Description{One hundred percent stacked bars, one per model and effort, splitting the 55 wrong patches into five coded outcomes, from rejected and named to accepted without naming the defect.}
\caption{Why Each Reviewer Rejected or Accepted the Wrong Patches. Each bar is one model and effort and splits the 55 wrong patches by the coded outcome of the review: rejected and named the tested defect, rejected and named it in part, rejected for another reason, accepted with feedback that named the defect, and accepted without naming it. Exploratory, post hoc coding by two language-model coders and an adjudicator, with no human validation. The bars carry no intervals; Table~\ref{tab:rq6} gives Wilson intervals for the named share. More effort moved wrong patches from accepted to rejected-and-named in every model, while rejections for another reason stayed at 1 to 7 of 55. Source: \href{https://github.com/trustmybot/research-artifacts/blob/cr-effort-claude-5.5-v1.1/papers/cr-effort-claude-5.5/findings/artifact/outputs/findings.json}{\code{findings.json}} (\code{by\_model.<model>.<effort>.patch\_outcomes}).}\label{fig:right-reason}
\end{figure*}

Most rejections of wrong patches named the defect that the failing strengthened test checks (Table~\ref{tab:rq6}, Figure~\ref{fig:right-reason}). Across the nine settings, 146 of the 205 rejections named that defect, 23 named it in part and 36 named something else; every rejection, malformed ones included, had at least one codable claim. Within a setting, the named share of rejections ran from .60 (Haiku at medium) to .82 (Sonnet at medium), with intervals that overlap for every pair of settings. As a share of all 55 wrong patches it rose with effort in every model: from .22 at low to .40 at high for Opus, from .24 to .40 for Sonnet and from .16 to .31 for Haiku. The coding was planned and run after the results were known, by language-model coders, and is exploratory.

\begin{table*}[t]
\caption{How Each Setting Handled the 55 Wrong Patches, by Coded Reason. Exploratory, post hoc coding by two language-model coders and an adjudicator. The coding is exploratory for every model. \textit{Rejected} counts fails and malformed verdicts on the 55 wrong patches. \textit{Named}: some finding states the same underlying issue as a reference behavior; \textit{partial}: the right condition or code region, but no consequence; \textit{other}: anything else. \textit{Feedback named the defect} counts accepted reviews whose feedback stated a reference behavior; \textit{specific concern, passed} counts accepted reviews that named a specific functional concern yet passed the patch. Effort raised the named share of the wrong patches in every model. Source: \href{https://github.com/trustmybot/research-artifacts/blob/cr-effort-claude-5.5-v1.1/papers/cr-effort-claude-5.5/findings/artifact/outputs/findings.json}{\code{findings.json}} (\code{by\_model.<model>.<effort>.right\_reason}, \code{accepted}).}\label{tab:rq6}
\footnotesize
\setlength{\tabcolsep}{4pt}
\begin{tabular}{@{}ll r c c r r r r r@{}}
\toprule
Model & Effort & Rejected & Named, of 55 [95\% CI] & Named, of rejections [95\% CI] & Partial & Other & Accepted & Feedback named the defect & Specific concern, passed \\
\midrule
Opus & Low & 18 & .22 [.13, .34] & .67 [.44, .84] & 2 & 4 & 37 & 11 & 22 \\
Opus & Medium & 29 & .38 [.27, .51] & .72 [.54, .85] & 3 & 5 & 26 & 12 & 16 \\
Opus & High & 30 & .40 [.28, .53] & .73 [.56, .86] & 3 & 5 & 25 & 14 & 16 \\
\addlinespace
Sonnet & Low & 19 & .24 [.14, .36] & .68 [.46, .85] & 3 & 3 & 36 & 12 & 19 \\
Sonnet & Medium & 22 & .33 [.22, .46] & .82 [.61, .93] & 3 & 1 & 33 & 9 & 15 \\
Sonnet & High & 28 & .40 [.28, .53] & .79 [.60, .90] & 2 & 4 & 27 & 12 & 18 \\
\addlinespace
Haiku & Low & 12 & .16 [.09, .28] & .75 [.47, .91] & 1 & 2 & 43 & 12 & 19 \\
Haiku & Medium & 20 & .22 [.13, .34] & .60 [.39, .78] & 1 & 7 & 35 & 13 & 20 \\
Haiku & High & 27 & .31 [.20, .44] & .63 [.44, .78] & 5 & 5 & 28 & 16 & 18 \\
\bottomrule
\end{tabular}
\end{table*}

\textbf{Effort steps.} Following each wrong patch from one effort to the next, the step from low to medium gained 9 named rejections and lost none for Opus, gained 7 and lost 2 for Sonnet, and gained 3 and lost none for Haiku; the step from medium to high gained 4 and lost 3 for Opus, 6 and 2 for Sonnet, and 7 and 2 for Haiku (\code{effort\_steps}). This mirrors RQ3: Opus gained mostly by medium, and Haiku more from medium to high. In Haiku's step from low to medium, rejections that named something else rose from 2 to 7, named ones from 9 to 12.

\textbf{Accepted reviews often described the defect.} Of the 290 accepted reviews of wrong patches, 111 named a reference behavior in their feedback and 64 more named one in part; 163 named a specific functional concern but passed the patch, 21 only hedged and 106 raised no concern. Haiku at high, for example, accepted 28 wrong patches, and 16 of those reviews named the defect. Passing a patch while describing its defect is a different failure from never noticing it, and it bears on RQ4: such a reviewer may have judged the tested behavior not required by the issue. We did not code which explanation applies.

\textbf{Findings were mostly functional.} The 232 rejecting reviews, false alarms included, carried 293 findings; 260 claimed a functional defect, 129 of them a logic defect (Table~\ref{tab:types}). The models differed most on larger defects, such as missing functionality (9, 10 and 1 for Opus, Sonnet and Haiku), and on requests for tests (2, 7 and 10). Of the 232 rejecting reviews, 216 showed a reproduction: a concrete input and its output, or a command the reviewer ran and its result.

\begin{table}[t]
\caption{Finding Types by Model, Pooled Over Effort. Types follow M\"antyl\"a and Lassenius \cite{mantyla2009defects}; a finding with several claims is coded by its first claim. Exploratory, post hoc coding. Logic defects led in every model. Source: \href{https://github.com/trustmybot/research-artifacts/blob/cr-effort-claude-5.5-v1.1/papers/cr-effort-claude-5.5/findings/artifact/outputs/findings.json}{\code{findings.json}} (\code{by\_model.<model>.<effort>.finding\_types}, \code{reproduction}). Subtypes are in supplement Section I.7.}\label{tab:types}
\footnotesize
\begin{tabular}{@{}l r r r r@{}}
\toprule
Type & Opus & Sonnet & Haiku & All \\
\midrule
Functional: logic & 44 & 46 & 39 & 129 \\
Functional: interface & 15 & 11 & 18 & 44 \\
Functional: check & 16 & 11 & 13 & 40 \\
Functional: resource & 7 & 10 & 9 & 26 \\
Functional: larger defects & 9 & 10 & 1 & 20 \\
Functional: support & 0 & 1 & 0 & 1 \\
Evolvability, all types & 3 & 1 & 5 & 9 \\
Not a defect: test request & 2 & 7 & 10 & 19 \\
Not a defect: other & 1 & 1 & 3 & 5 \\
\midrule
Findings & 97 & 98 & 98 & 293 \\
Rejecting reviews & 83 & 77 & 72 & 232 \\
With a reproduction & 79 & 72 & 65 & 216 \\
\bottomrule
\end{tabular}
\end{table}

\textbf{False alarms mostly flagged behavior the official fix shares.} Of the 27 rejections of correct patches, 18 claimed a real defect that the official fix shares, 5 a real defect outside what the issue asks, 1 a real defect within the issue and 1 a defect that does not occur; 2 could not be verified from the material. Seventeen were coded as beyond the objective of the change \cite{lin2026coderabbit}, 15 as demanding more than the issue asks \cite{jin2026overcorrection} and 1 as a misidentified issue. Haiku at medium gave 7 of the 27 and every other setting 2 to 4. By the label rule that made the patch correct (supplement Section A.5), 16 fell on patches that behave like the official fix, 9 on patches that share its defect and 2 on patches whose failing test checks only an implementation detail. Agreement on the beyond-the-objective code was low before adjudication ($\kappa = .20$, 95\% CI [.01, .42]; supplement Table S22), so that count is the least certain here.

\textbf{Decoy cross-check.} Of the 27 decoy rejections of wrong patches, the claim coders judged 24 to report a real defect within the issue; the match coders had scored 18 of those 24 as named, 1 as partial and 5 as other. A rejection scored as naming something else can therefore still report a real defect that the failing test does not check, so the named share is a lower bound on rejections for a real reason.

\textbf{Sensitivity.} The named shares change little when a partial match counts as named, when the patch with a contested reference behavior is dropped (no share moves by more than .01) or when malformed rejections are dropped (supplement Table S25). Without the five patches shown wrong by reading, Opus no longer rises from medium to high (20 and then 19 of 50).

\section{Discussion}\label{sec:discussion}

\subsection{Answers to the Research Questions}\label{sec:answers}

For Claude 5.5 models reviewing Django patches with tools and the issue text, more effort than low helped in every model, and more than medium helped little for Opus but substantially for Sonnet and Haiku. High accepted fewer plausible-but-wrong patches than low in all three models, a supported difference for Opus (confirmatory) and Haiku (exploratory) but not for Sonnet at the exact test's threshold (RQ1). Medium matched high within the margin for Opus, with near-identical verdicts, at a saving below what we had preregistered as material; for Sonnet and Haiku, medium saved a material share of the cost but accepted enough more wrong patches that non-inferiority is not shown (RQ2). The step from low to medium bought most of Opus's gain, and the step from medium to high most of Sonnet's and Haiku's; under the original labels, Opus agreed more at medium than at high (RQ3). These conclusions depend on the reference standard (RQ4): in every model, most false successes fall on wrong patches that fail a strengthened test the issue may not require, and for Opus, recall of the upstream fix is common and goes with accepting wrong patches. Across models (RQ5), Opus at medium or high agreed best; Haiku at high came within an unresolved distance at about 8\% of the cost per review, and Sonnet at low or medium and Opus at low were beaten on both cost and agreement by Haiku at high. In exploratory coding (RQ6), most rejections of wrong patches named the defect the failing test checks (146 of 205 across settings), and effort raised the named share of the 55 wrong patches in every model, mostly by medium for Opus and more at high for Haiku; 111 of the 290 acceptances of wrong patches named the defect yet passed the patch, and most false alarms pointed at behavior the official fix shares.

\subsection{Interpretation}\label{sec:interpretation}

High does more work than medium in every model, with more output on more than 90\% of patches, but how much that work changes the verdict differs. For Opus it rarely did and did not add test runs; the gain from low to medium, by contrast, came with more tool calls and test runs. For Haiku, high roughly doubled the tool calls and turns of medium and changed 15 verdicts on wrong patches. The reviewer can both read and run code, the two kinds of verifier that R2E-Gym found complementary and strongest combined \cite{jain2025r2egym}, yet at low effort Opus ran no shell test in 23 of 90 reviews and Sonnet in 34 (Figure~\ref{fig:flowchart}). We did not analyze mediation, so this describes co-occurring behavior, not a mechanism. The pattern, effort that helps until a point that differs by model, matches reports that reasoning judges beat non-reasoning ones \cite{jiang2026codejudgebench} while gains from added effort vary by model \cite{huang2025judge}.

The return on effort leveled off by medium for Opus only. From medium to high, Opus caught 1 more wrong patch, Sonnet 6 and Haiku 7; from low to medium, Opus caught 11, Sonnet 3 and Haiku 8 (Table~\ref{tab:results}). The gain was not proportional to model strength: from low to high, Haiku gained more catches (15) than Opus (12). With three models of one family, this is a pattern of this experiment, not an estimate of how the gain scales with capability.

The two errors were lopsided. In all nine settings the reviewer accepted far more wrong patches than it rejected correct ones: false success ran from .45 to .78 and false alarm from .06 to .20 (Table~\ref{tab:results}, Figure~\ref{fig:error-rates}). Either the reviewer misses real defects the issue implies, or the tests demand more than the issue says and the reviewer's acceptance is defensible. The issue-requirement coding favors the second explanation for many patches in every model (Table~\ref{tab:requires}), but one post hoc model coder cannot settle it. The low detection rate is consistent with the adversarial-patch results of \citet{trad2026adversarial}, in a harder setting, and the role of strict tests connects to work on weak and overfit benchmark tests \cite{ahmed2026overfitting,wang2026solved,yu2026sweabs}. The zero flip rate for Opus contrasts with reports of run-to-run disagreement in judges \cite{zhao2026bias}, although our bound is loose.

Memory of the fix did not make the reviewer better. One might expect a model that has seen the upstream fix in training, as SWE-bench models often have \cite{liang2025illusion}, to spot a patch that departs from it. For Opus, the opposite held: wrong patches on issues whose fix it fully recalled were accepted more often at every effort than those on partially recalled issues (.79 against .54 at low, .64 against .27 at medium and .57 against .31 at high; Figure~\ref{fig:recall-split}), and reviews flagged as drawing on memory of the fix accepted 71\% of wrong patches against 46\% for unflagged ones. A reviewer that recognizes the issue may judge the patch against the fix it remembers, and accept a patch that resembles that fix, rather than check it against the issue and the code. This evidence is exploratory, correlational and for Opus only: issues whose fixes the model recalls may also be harder to review, and we did not measure recall for Sonnet or Haiku. False success stayed high in every model and setting, from .45 to .78, so no model reached a strong gate however much of the codebase it may have seen in training.

\textbf{Seen but accepted.} The RQ6 coding adds a second view of false success. Of the 290 accepted reviews of wrong patches, 111 named the tested defect in their feedback, and more than half named a specific concern while passing the patch. For these reviews the failure lay not in finding the defect but in deciding that it mattered, which fits the RQ4 reading that the tested behavior may lie beyond what the issue asks. Effort may therefore work partly by changing that decision rather than by finding more; the coding cannot separate the two, and it is exploratory.

\subsection{Implications}\label{sec:implications}

For practitioners using Opus 5.5 as a review gate under similar conditions, medium is a sound default: it costs less than high and gave almost the same verdicts. This matches the vendor's default for Opus 5.5 and its advice to set effort from a sweep on one's own evaluations \cite{anthropic2026effort}. The default does not carry over to smaller models. With Sonnet and Haiku, high caught clearly more wrong patches than medium, and with Haiku it still cost under a cent a review. The vendor suggests starting Haiku at medium for most work, including agentic coding, and moving to high for longer agent tasks \cite{anthropic2026effort}; for this review gate our data point to high. A team that must cut cost could consider Haiku at high, at about 8\% of Opus medium's cost per review, but its equivalence to Opus at medium is untested, it had more false alarms, and its malformed verdicts need handling. That comparison is exploratory and rests on one run per review, so it motivates a confirmatory test rather than a switch. No setting is a strong gate against plausible-but-wrong patches; a review gate of this kind complements tests rather than replacing them.

For researchers, benchmark labels from strengthened tests need a human check of whether each failing test reflects the issue before they can serve as ground truth for reviewers, and contamination probes should use several runs per issue, since one run misgroups some issues. Balanced agreement, our summary measure, weights a false success and a false alarm equally. For a review gate, an accepted wrong patch usually costs more than a rejected correct one, and the errors here were lopsided. Future studies should therefore preregister a cost-weighted measure, such as an F-score that weights missed wrong patches more heavily, with the weight and the expected share of wrong patches stated in advance. Measures over all decision thresholds, such as the area under the ROC curve, would need the reviewer to report a graded confidence rather than only pass or fail.

\section{Threats to Validity}\label{sec:threats}

We group threats following \citet{wohlin2012experimentation} and pair each with what the study did about it.

\textbf{Construct validity.} \textit{The labels are not ground truth.} SWE-ABS tests derive from the official fix and are partly model-generated. The labels were corrected after data collection by written rules that language-model agents applied, with changes proposed by the agents and the co-authors and each ruled on by the first author, but no independent rater checked them, and correcting labels after the verdicts were known risks moving them toward the reviewer, the bias of discrepant analysis \cite{hadgu1999discrepant}. We checked every label, not only discrepant ones, by rules fixed before rescoring; the correction moved labels both ways; and the original-label results give the same hypothesis decisions in every model (Section~\ref{sec:robust}). The label-error sensitivity analysis bounds random label error but not systematic strictness, which the issue-requirement coding can flag but not measure, since it was post hoc and by one model coder. \textit{What a pass means.} The prompt fails a patch only for functional problems, so the study says nothing about the quality of security, performance or design feedback. \textit{Effort delivery.} The CLI does not report the effort applied; delivery rests on a request capture before the run and on the behavioral differences between efforts. \textit{Equal error weights.} Balanced agreement treats both errors as equally costly, which a review gate may not (Section~\ref{sec:implications}).

\textbf{Internal validity.} \textit{Infrastructure change.} The machine type and harness version changed after 12 Opus patches, in patch order (supplement Table S3, row 1); an effect through, for example, tool latency cannot be ruled out. \textit{Order and drift.} For 32 Opus reviews the efforts of a patch did not run back to back (Table~\ref{tab:deviations}, row 1), and the model identifier carries no date. \textit{Peeking.} Cost and token data were visible during the internal Opus pilot, whose decision was not recorded (supplement Table S3, row 3); verdicts by effort stayed closed until the batch ended. \textit{Contamination.} All Verified issues predate 2024, and Opus recalls most of their fixes; with no internet access, that recall can come only from training. The paired design removes an effect that shifts all efforts equally, but not one that differs by effort; flagged reviews were somewhat fewer at high. Recall was not measured for Sonnet or Haiku. \textit{Model coders.} The memory flags and issue-requirement codes are judgments of Claude Opus 5.5, the model under test, without human validation, so the coders may share the reviewer's blind spots. \textit{Shallow passes.} Some passes, mostly at low effort, came with no thinking, at most one tool call or no test run (Figure~\ref{fig:flowchart}), so they may be acceptances by default rather than checked judgments. Language-model judges recognize valid outputs far better than invalid ones \cite{jain2025agreeableness}, and superficial features of code can push a judge toward a ``correct'' verdict \cite{moon2026cover}. Such passes raise agreement on correct patches without earning it, so low effort's few false alarms likely overstate how well it checks; we did not read or re-score these reviews. \textit{Cross-model comparison.} Sonnet and Haiku ran three days after Opus at a later harness version, not interleaved with it (Table~\ref{tab:deviations}, row 11), so drift between the runs is confounded with the model. An effort level need not mean the same amount of work in every model: the defaults differ (high for Sonnet 5.5, medium for Opus 5.5 and Haiku 5.5), and the vendor notes that a level's calibration can change between models \cite{anthropic2026effort}, so the comparisons are between configurations, not between models at equal effort.

\textbf{External validity.} The confirmatory study covers one model, the exploratory studies two more of the same family, and all of them three of five effort levels, one repository in one language, one prompt, one route and public agent patches. Richer reviewer context, such as a code dependency graph or project review rules, may change the effect of effort. Reviews of new private code would not benefit from recall of a known fix, so absolute rates may differ there.

\textbf{Conclusion validity.} With 35 correct patches, false-alarm intervals are wide and the guardrail and ceiling are coarse; both were set for 30 correct patches and kept unchanged, which makes them slightly more lenient as shares. The exploratory subgroups are small (23 and 32 patches) and uncorrected for multiplicity, so they are not tests; only the three Opus tests are confirmatory. The Holm correction within each smaller model controls error only inside that model, not across the six exploratory tests, and the comparison of nine settings in RQ5 is descriptive. Twenty repeats bound the flip rate only to below about 16\%, and only for Opus; the smaller models' run-to-run noise is unmeasured.

\textbf{RQ6 coding.} \textit{Construct.} The named code asks whether a review states the behavior the failing strengthened test checks, so it inherits the limits of those tests (Section~\ref{sec:requires}); a rejection coded as naming something else may still report a real defect, as the decoy check shows. For the five patches shown wrong by reading, the reference behaviors come from that reading, not from a failing test; dropping them changes no named share by more than .02 but removes Opus's rise from medium to high. One reference behavior rests on a strengthened test that the checking agent judged overfitted to an internal function signature of the official fix; the sensitivity analysis drops the patch that carries it, and no named share moves by more than .01. \textit{Internal.} RQ6 and its codebook were written after all results above were known. Both coders and the adjudicator are Claude 5.5 models, the family under test, and language-model evaluators can favor outputs like their own \cite{panickssery2024selfpref}. The coders were not shown the model or effort; coder A guessed the model correctly no more often than chance (69 of 232, .30, 95\% CI [.24, .36], against .33), and coder B declined to guess every time. The coders of matches and of accepted reviews knew that every item was a wrong patch. The coders read their instructions and wrote their codes with file tools, although the prompts asked for coding without tools, and tool use was audited only for the runs that drafted the reference behaviors. Three items lacked one coder's code and were not rerun, and coder disagreements the codebook does not send to adjudication took the lower score, which changed no review-level code (supplement Section I.5). \textit{External.} The coding covers only this study's reviews of one repository. \textit{Conclusion.} Each setting has 12 to 30 rejections, so the intervals are wide, and no difference between settings was tested. \textit{No human validation.} No person checked the reference behaviors, the codes or the adjudications; agreement between two model coders shows consistency, not accuracy.

\section{Conclusion}\label{sec:conclusion}

In this paired experiment, three Claude 5.5 models reviewing Django patches accepted fewer plausible-but-wrong patches at high effort than at low. For Opus 5.5, in the preregistered study, medium matched high at lower cost, though its saving fell short of what we had preregistered as material. For Sonnet 5.5 and Haiku 5.5, in exploratory studies, medium was much cheaper than high but not shown to be as accurate, so the gains from effort leveled off at medium for Opus only. These decisions hold under both the corrected and the original labels. Opus at medium or high agreed best with the labels; Haiku at high cost a small fraction as much and came within a distance the data cannot resolve. Every setting accepted far more wrong patches than it rejected correct ones, and exploratory analyses suggest that strict strengthened tests, in every model, explain part of that. Memory of the fix did not help: for Opus, the only model we tested for recall, wrong patches on issues whose fix it fully recalled were accepted more often at every effort, not less. This is an exploratory association, and in every model false success stayed between .45 and .78 whatever the model may have seen in training. Exploratory coding of the reviews suggests that most rejections named the defect the strengthened test checks, that effort added such rejections, and that many accepting reviews described the defect but passed the patch anyway.

Future work follows directly. First, people should code whether each issue requires the behavior its failing strengthened test checks, and an independent audit should check the corrected labels, starting with the patches on which every effort disagreed with the label and including agreeing patches. Second, recall should be measured with several runs per issue and for every model. Third, a preregistered study with repeated runs should test whether Haiku 5.5 at high is equivalent to Opus 5.5 at medium, scored by a cost-weighted measure fixed in advance that counts an accepted wrong patch as worse than a rejected correct one. Fourth, richer reviewer conditions, such as review checklists, project rules or a code dependency graph, should be compared with effort; the later studies in this series take up the author's note, reviewer skills and voting. Finally, new private code, where no fix can be recalled, would show how much the absolute rates owe to memory.

\section{Data Availability}\label{sec:data}

The data, analysis code and figures are in the public trustmybot/\allowbreak research-artifacts repository at tag cr-effort-claude-5.5-v1.2, at \url{https://github.com/trustmybot/research-artifacts/tree/cr-effort-claude-5.5-v1.2/papers/cr-effort-claude-5.5}; the code is under the MIT License and the data, text and figures under CC BY 4.0. The artifact directory (\url{https://github.com/trustmybot/research-artifacts/tree/cr-effort-claude-5.5-v1.2/papers/cr-effort-claude-5.5/artifact}) holds one row per Opus review (270 main reviews and 20 repeats), every model-call attempt, the spend record, the memory coding, both recall runs, the label checks, the issue-requirement coding, and the original and corrected label of every patch with the rule that decided it (\url{https://github.com/trustmybot/research-artifacts/blob/cr-effort-claude-5.5-v1.0/papers/cr-effort-claude-5.5/artifact/data/labels.csv}). The commands in \url{https://github.com/trustmybot/research-artifacts/blob/cr-effort-claude-5.5-v1.2/papers/cr-effort-claude-5.5/artifact/README.md} reproduce \url{https://github.com/trustmybot/research-artifacts/blob/cr-effort-claude-5.5-v1.2/papers/cr-effort-claude-5.5/artifact/outputs/results.json}, the source of the Opus-only analyses; they need only Node.js 24, and the bootstrap is seeded, so reruns are identical. The reviews of all three models are in \url{https://github.com/trustmybot/research-artifacts/blob/cr-effort-claude-5.5-v1.0/papers/cr-effort-claude-5.5/haiku-vs-opus/artifact/data/reviews.csv}, one row per review with no transcript text; the same README gives the commands that reproduce the all-model results (\url{https://github.com/trustmybot/research-artifacts/blob/cr-effort-claude-5.5-v1.2/papers/cr-effort-claude-5.5/artifact/outputs/models.json}), the swap of the three replaced patches (\url{https://github.com/trustmybot/research-artifacts/blob/cr-effort-claude-5.5-v1.2/papers/cr-effort-claude-5.5/artifact/outputs/models-swap.json}) and the shallow-pass counts (\url{https://github.com/trustmybot/research-artifacts/blob/cr-effort-claude-5.5-v1.0/papers/cr-effort-claude-5.5/artifact/outputs/shallow-pass.json}). The supplement (\url{https://github.com/trustmybot/research-artifacts/blob/cr-effort-claude-5.5-v1.2/papers/cr-effort-claude-5.5/supplement.md}) gives the full procedure (Section A), every analysis and statistical test (Section B), the deviations in full (Section C), the Opus tables (Section D), the spend record (Section E), wall time by machine type (Section F), the Sonnet and Haiku tables and spend (Section G) and the shallow passes (Section H). The exploratory note comparing the catches of Haiku at high and Opus at medium patch by patch is \url{https://github.com/trustmybot/research-artifacts/blob/cr-effort-claude-5.5-v1.2/papers/cr-effort-claude-5.5/haiku-vs-opus/analysis.md}. Full transcripts contain patch text with no license for redistribution and are kept private; they are available from the authors on request, and the artifact README describes how to rebuild the data from them. The RQ6 coding is in the findings directory (\url{https://github.com/trustmybot/research-artifacts/tree/cr-effort-claude-5.5-v1.1/papers/cr-effort-claude-5.5/findings}), with its codebook, the prompts for the reference behaviors, coders and adjudicator, the analysis code, every aggregate the paper reports (\href{https://github.com/trustmybot/research-artifacts/blob/cr-effort-claude-5.5-v1.1/papers/cr-effort-claude-5.5/findings/artifact/outputs/findings.json}{\code{findings.json}}) and Figure~\ref{fig:right-reason}. The 57 reference behaviors of the 55 wrong patches are released with patch code elided (\href{https://github.com/trustmybot/research-artifacts/blob/cr-effort-claude-5.5-v1.1/papers/cr-effort-claude-5.5/findings/artifact/data/gold-behaviours.json}{\code{gold-behaviours.json}}). Also released are every reviewer finding with patch code elided (\href{https://github.com/trustmybot/research-artifacts/blob/cr-effort-claude-5.5-v1.1/papers/cr-effort-claude-5.5/findings/artifact/data/findings.csv}{\code{findings.csv}}), every code the adjudicator decided, with both coders' codes, the final code and the codebook rule cited (\href{https://github.com/trustmybot/research-artifacts/blob/cr-effort-claude-5.5-v1.1/papers/cr-effort-claude-5.5/findings/artifact/data/adjudication-log.csv}{\code{adjudication-log.csv}}), and the key from each blinded item to its review (\href{https://github.com/trustmybot/research-artifacts/blob/cr-effort-claude-5.5-v1.1/papers/cr-effort-claude-5.5/findings/artifact/data/blind-key.json}{\code{blind-key.json}}). The settled per-item codes, the agreement records and the coder runs are kept private, so findings.json cannot yet be recomputed from public data (supplement Section I.11).

\begin{acks}
\textbf{Use of AI.} An AI agent (Claude Opus 5.5, in the researcher role of the Trust My Bot agent harness) wrote the analysis scripts, computed every statistic, drew the figures and drafted this manuscript. Claude Opus 5.5, one of the models under test, also did the memory coding and the issue-requirement coding. For RQ6, Claude Opus 5.5 drafted the reference behaviors and Claude Sonnet 5.5 checked them; Claude Opus 5.5 and Claude Sonnet 5.5 coded the reviews, and Claude Opus 5.5 adjudicated their disagreements; no person checked these codes. Language-model agents applied the written label rules and the three authors each checked 30 of the 90 patches; both the agents and the co-authors proposed label changes, and the first author ruled on each one. The authors reviewed the analysis, the figures and the text, and take full responsibility for the content of this paper.

\textbf{Competing interests.} Zax Shen develops Trust My Bot, the agent harness whose review gate motivated this study. Hanling Zhang is a member of Trust My Bot. Zixu Wang declares no competing interests.

\textbf{Funding.} The study received no external funding. Zax Shen paid for the experiments; each author paid for their own independent review of the results. Model use was paid at list price; no research credits were used.
\end{acks}

\bibliographystyle{ACM-Reference-Format}
\bibliography{references}

@inproceedings{ahmed2026overfitting,
  author    = {Ahmed, Toufique and Ganhotra, Jatin and Shinnar, Avraham and Hirzel, Martin},
  title     = {Investigating test overfitting on {SWE-bench}},
  booktitle = {Proceedings of the 34th ACM International Conference on the Foundations of Software Engineering (FSE Companion '26)},
  publisher = {Association for Computing Machinery},
  address   = {New York, NY, USA},
  pages     = {1292--1296},
  year      = {2026},
  doi       = {10.1145/3803437.3805574}
}

@misc{anthropic2026effort,
  author       = {{Anthropic}},
  title        = {Effort},
  howpublished = {Claude Platform Docs},
  year         = {2026},
  url          = {https://platform.claude.com/docs/en/build-with-claude/effort},
  lastaccessed = {November 13, 2026}
}

@misc{anthropic2026model,
  author       = {{Anthropic}},
  title        = {Model configuration},
  howpublished = {Claude Code Docs},
  year         = {2026},
  url          = {https://code.claude.com/docs/en/model-config},
  lastaccessed = {October 27, 2026}
}

@misc{anthropic2026pricing,
  author       = {{Anthropic}},
  title        = {Pricing},
  howpublished = {Claude Platform Docs},
  year         = {2026},
  url          = {https://platform.claude.com/docs/en/about-claude/pricing},
  lastaccessed = {October 27, 2026}
}

@misc{chowdhury2024verified,
  author       = {Chowdhury, Neil and Aung, James and Shern, Chan Jun and Jaffe, Oliver and Sherburn, Dane and Starace, Giulio and Mays, Evan and Dias, Rachel and Aljubeh, Marwan and Glaese, Mia and Jimenez, Carlos E. and Yang, John and Ho, Leyton and Patwardhan, Tejal and Liu, Kevin and Madry, Aleksander},
  title        = {Introducing {SWE-bench Verified}},
  howpublished = {OpenAI},
  year         = {2024},
  url          = {https://openai.com/index/introducing-swe-bench-verified/}
}

@article{cohen1960kappa,
  author  = {Cohen, Jacob},
  title   = {A coefficient of agreement for nominal scales},
  journal = {Educational and Psychological Measurement},
  volume  = {20},
  number  = {1},
  pages   = {37--46},
  year    = {1960},
  doi     = {10.1177/001316446002000104}
}

@article{efron1987bca,
  author  = {Efron, Bradley},
  title   = {Better bootstrap confidence intervals},
  journal = {Journal of the American Statistical Association},
  volume  = {82},
  number  = {397},
  pages   = {171--185},
  year    = {1987},
  doi     = {10.1080/01621459.1987.10478410}
}

@article{fagerland2013mcnemar,
  author    = {Fagerland, Morten W. and Lydersen, Stian and Laake, Petter},
  title     = {The {McNemar} test for binary matched-pairs data: {Mid-p} and asymptotic are better than exact conditional},
  journal   = {BMC Medical Research Methodology},
  volume    = {13},
  articleno = {91},
  year      = {2013},
  doi       = {10.1186/1471-2288-13-91}
}

@misc{farosai2025paradox,
  author       = {{Faros AI}},
  title        = {The {AI} productivity paradox report 2025},
  howpublished = {Industry report},
  year         = {2025},
  url          = {https://www.faros.ai/blog/ai-productivity-paradox-report-2025},
  lastaccessed = {November 17, 2026}
}

@article{hadgu1999discrepant,
  author  = {Hadgu, Alula},
  title   = {Discrepant analysis: {A} biased and an unscientific method for estimating test sensitivity and specificity},
  journal = {Journal of Clinical Epidemiology},
  volume  = {52},
  number  = {12},
  pages   = {1231--1237},
  year    = {1999},
  doi     = {10.1016/S0895-4356(99)00101-8}
}

@inproceedings{he2026cursor,
  author    = {He, Hao and Miller, Courtney and Agarwal, Shyam and K{\"a}stner, Christian and Vasilescu, Bogdan},
  title     = {Speed at the cost of quality: {How} {Cursor} {AI} increases short-term velocity and long-term complexity in open-source projects},
  booktitle = {Proceedings of the 23rd International Conference on Mining Software Repositories (MSR '26)},
  publisher = {Association for Computing Machinery},
  address   = {New York, NY, USA},
  year      = {2026},
  doi       = {10.1145/3793302.3793349}
}

@article{holm1979sequential,
  author  = {Holm, Sture},
  title   = {A simple sequentially rejective multiple test procedure},
  journal = {Scandinavian Journal of Statistics},
  volume  = {6},
  number  = {2},
  pages   = {65--70},
  year    = {1979},
  url     = {https://www.jstor.org/stable/4615733}
}

@misc{huang2025judge,
  author        = {Huang, Donghao and Chew, Shila and Dutkiewicz, Anna and Wang, Zhaoxia},
  title         = {{LLM-as-a-judge} for scalable test coverage evaluation: {Accuracy}, operational reliability, and cost},
  year          = {2025},
  archiveprefix = {arXiv},
  eprint        = {2512.01232},
  doi           = {10.48550/arXiv.2512.01232}
}

@article{imai2010mediation,
  author  = {Imai, Kosuke and Keele, Luke and Yamamoto, Teppei},
  title   = {Identification, inference and sensitivity analysis for causal mediation effects},
  journal = {Statistical Science},
  volume  = {25},
  number  = {1},
  pages   = {51--71},
  year    = {2010},
  doi     = {10.1214/10-STS321}
}

@inproceedings{jain2025r2egym,
  author    = {Jain, Naman and Singh, Jaskirat and Shetty, Manish and Zhang, Tianjun and Zheng, Liang and Sen, Koushik and Stoica, Ion},
  title     = {{R2E-Gym}: {Procedural} environment generation and hybrid verifiers for scaling open-weights {SWE} agents},
  booktitle = {Second Conference on Language Modeling (COLM 2025)},
  year      = {2025},
  url       = {https://openreview.net/forum?id=7evvwwdo3z}
}

@misc{jain2025agreeableness,
  author        = {Jain, Suryaansh and Ahmed, Umair Z. and Sahai, Shubham and Leong, Ben},
  title         = {Beyond consensus: {Mitigating} the agreeableness bias in {LLM} judge evaluations},
  year          = {2025},
  archiveprefix = {arXiv},
  eprint        = {2510.11822},
  doi           = {10.48550/arXiv.2510.11822}
}

@inproceedings{jiang2026codejudgebench,
  author    = {Jiang, Hongchao and Chen, Yiming and Cao, Yushi and Lee, Hung-yi and Tan, Robby T.},
  title     = {{CodeJudgeBench}: {Benchmarking} {LLM-as-a-judge} for coding tasks},
  booktitle = {Proceedings of the 64th Annual Meeting of the Association for Computational Linguistics (Volume 1: Long Papers)},
  publisher = {Association for Computational Linguistics},
  pages     = {19416--19448},
  year      = {2026},
  doi       = {10.18653/v1/2026.acl-long.888}
}

@inproceedings{jimenez2024swebench,
  author    = {Jimenez, Carlos E. and Yang, John and Wettig, Alexander and Yao, Shunyu and Pei, Kexin and Press, Ofir and Narasimhan, Karthik},
  title     = {{SWE-bench}: {Can} language models resolve real-world {GitHub} issues?},
  booktitle = {The Twelfth International Conference on Learning Representations (ICLR 2024)},
  year      = {2024},
  url       = {https://openreview.net/forum?id=VTF8yNQM66}
}

@misc{li2025teammates,
  author        = {Li, Hao and Zhang, Haoxiang and Hassan, Ahmed E.},
  title         = {The rise of {AI} teammates in software engineering {(SE)} 3.0: {How} autonomous coding agents are reshaping software engineering},
  year          = {2025},
  archiveprefix = {arXiv},
  eprint        = {2507.15003v1},
  doi           = {10.48550/arXiv.2507.15003}
}

@misc{liang2025illusion,
  author        = {Liang, Shanchao and Garg, Spandan and Moghaddam, Roshanak Zilouchian},
  title         = {The {SWE-bench} illusion: {When} state-of-the-art {LLMs} remember instead of reason},
  year          = {2025},
  archiveprefix = {arXiv},
  eprint        = {2506.12286v4},
  doi           = {10.48550/arXiv.2506.12286}
}

@article{mcnemar1947,
  author  = {McNemar, Quinn},
  title   = {Note on the sampling error of the difference between correlated proportions or percentages},
  journal = {Psychometrika},
  volume  = {12},
  number  = {2},
  pages   = {153--157},
  year    = {1947},
  doi     = {10.1007/BF02295996}
}

@inproceedings{moon2026cover,
  author    = {Moon, Jiwon and Hwang, Yerin and Lee, Dongryeol and Kang, Taegwan and Kim, Yongil and Jung, Kyomin},
  title     = {Don't judge code by its cover: {Exploring} biases in {LLM} judges for code evaluation},
  booktitle = {Findings of the Association for Computational Linguistics: EACL 2026},
  publisher = {Association for Computational Linguistics},
  pages     = {1364--1389},
  year      = {2026},
  doi       = {10.18653/v1/2026.findings-eacl.70}
}

@article{newcombe1998paired,
  author  = {Newcombe, Robert G.},
  title   = {Improved confidence intervals for the difference between binomial proportions based on paired data},
  journal = {Statistics in Medicine},
  volume  = {17},
  number  = {22},
  pages   = {2635--2650},
  year    = {1998},
  doi     = {10.1002/(SICI)1097-0258(19981130)17:22<2635::AID-SIM954>3.0.CO;2-C}
}

@misc{openagenteval2026sweabs,
  author       = {{OpenAgentEval}},
  title        = {{SWE-ABS}},
  howpublished = {Computer software and data set, GitHub},
  year         = {2026},
  url          = {https://github.com/OpenAgentEval/SWE-ABS}
}

@inproceedings{shi2026hitl,
  author    = {Shi, Sherry and Wei, Renyao and Tufano, Michele and Cambronero, Jos{\'e} and Cheng, Runxiang and Ivan{\v{c}}i{\'c}, Franjo and Rondon, Pat},
  title     = {Towards a human-in-the-loop framework for reliable patch evaluation using an {LLM-as-a-judge}},
  booktitle = {Proceedings of the 1st International Workshop on AI for Software Quality Evaluation: Judgment, Metrics, Benchmarks, and Beyond (AI-SQE '26)},
  publisher = {Association for Computing Machinery},
  address   = {New York, NY, USA},
  pages     = {19--28},
  year      = {2026},
  doi       = {10.1145/3786175.3788348}
}

@misc{swebench2026experiments,
  author       = {{SWE-bench}},
  title        = {experiments},
  howpublished = {Data set, GitHub},
  year         = {2026},
  url          = {https://github.com/swe-bench/experiments}
}

@article{tango1998equivalence,
  author  = {Tango, Toshiro},
  title   = {Equivalence test and confidence interval for the difference in proportions for the paired-sample design},
  journal = {Statistics in Medicine},
  volume  = {17},
  number  = {8},
  pages   = {891--908},
  year    = {1998},
  doi     = {10.1002/(SICI)1097-0258(19980430)17:8<891::AID-SIM780>3.0.CO;2-B}
}

@misc{toon,
  author       = {{TOON Format}},
  title        = {{\texttt{@toon-format/toon}}},
  howpublished = {Computer software, GitHub},
  year         = {2025},
  url          = {https://github.com/toon-format/toon}
}

@misc{trad2026adversarial,
  author        = {Trad, Fares and Chen, Simin and Pham, Hung Viet and Uddin, Gias and Ray, Baishakhi},
  title         = {Adversarial testing of automated program repair agents for security vulnerabilities},
  year          = {2026},
  archiveprefix = {arXiv},
  eprint        = {2609.15963},
  doi           = {10.48550/arXiv.2609.15963}
}

@inproceedings{wang2026solved,
  author    = {Wang, You and Pradel, Michael and Liu, Zhongxin},
  title     = {Are ``solved issues'' in {SWE-bench} really solved correctly? {An} empirical study},
  booktitle = {Proceedings of the 2026 IEEE/ACM 48th International Conference on Software Engineering (ICSE '26)},
  publisher = {Association for Computing Machinery},
  address   = {New York, NY, USA},
  pages     = {169--181},
  year      = {2026},
  doi       = {10.1145/3744916.3764576}
}

@article{wilcoxon1945,
  author  = {Wilcoxon, Frank},
  title   = {Individual comparisons by ranking methods},
  journal = {Biometrics Bulletin},
  volume  = {1},
  number  = {6},
  pages   = {80--83},
  year    = {1945},
  doi     = {10.2307/3001968}
}

@article{wilson1927,
  author  = {Wilson, Edwin B.},
  title   = {Probable inference, the law of succession, and statistical inference},
  journal = {Journal of the American Statistical Association},
  volume  = {22},
  number  = {158},
  pages   = {209--212},
  year    = {1927},
  doi     = {10.1080/01621459.1927.10502953}
}

@book{wohlin2012experimentation,
  author    = {Wohlin, Claes and Runeson, Per and H{\"o}st, Martin and Ohlsson, Magnus C. and Regnell, Bj{\"o}rn and Wessl{\'e}n, Anders},
  title     = {Experimentation in software engineering},
  publisher = {Springer},
  address   = {Berlin, Heidelberg},
  year      = {2012},
  doi       = {10.1007/978-3-642-29044-2}
}

@inproceedings{yu2026sweabs,
  author    = {Yu, Boxi and Cao, Yang and Zhang, Yuzhong and Lin, Liting and Xu, Junjielong and Zhong, Zhiqing and Xu, Qinghua and Wang, Guancheng and Cao, Jialun and Cheung, Shing-Chi and He, Pinjia and Briand, Lionel},
  title     = {{SWE-ABS}: {Adversarial} benchmark strengthening exposes inflated success rates on test-based benchmark},
  booktitle = {Proceedings of the 43rd International Conference on Machine Learning},
  series    = {Proceedings of Machine Learning Research},
  volume    = {306},
  publisher = {PMLR},
  pages     = {149974--150011},
  year      = {2026},
  url       = {https://proceedings.mlr.press/v306/yu26b.html}
}

@misc{zhao2026bias,
  author        = {Zhao, Zixiao and Esmaeili, Amirreza and Fard, Fatemeh},
  title         = {Bias in the loop: {Auditing} {LLM-as-a-judge} for software engineering},
  year          = {2026},
  archiveprefix = {arXiv},
  eprint        = {2604.16790},
  doi           = {10.48550/arXiv.2604.16790}
}

@inproceedings{zheng2023judging,
  author    = {Zheng, Lianmin and Chiang, Wei-Lin and Sheng, Ying and Zhuang, Siyuan and Wu, Zhanghao and Zhuang, Yonghao and Lin, Zi and Li, Zhuohan and Li, Dacheng and Xing, Eric P. and Zhang, Hao and Gonzalez, Joseph E. and Stoica, Ion},
  title     = {Judging {LLM-as-a-judge} with {MT-Bench} and {Chatbot Arena}},
  booktitle = {Advances in Neural Information Processing Systems 36 (NeurIPS 2023), Datasets and Benchmarks Track},
  pages     = {46595--46623},
  year      = {2023},
  doi       = {10.52202/075280-2020}
}

@article{gwet2008ac1,
  author  = {Gwet, Kilem Li},
  title   = {Computing inter-rater reliability and its variance in the presence of high agreement},
  journal = {British Journal of Mathematical and Statistical Psychology},
  volume  = {61},
  number  = {1},
  pages   = {29--48},
  year    = {2008},
  doi     = {10.1348/000711006X126600}
}

@misc{jin2026overcorrection,
  author        = {Jin, Haolin and Chen, Huaming},
  title         = {Are {LLMs} reliable code reviewers? {Systematic} overcorrection in requirement conformance judgement},
  year          = {2026},
  eprint        = {2603.00539},
  archivePrefix = {arXiv},
  doi           = {10.48550/arXiv.2603.00539}
}

@misc{lin2026coderabbit,
  author        = {Lin, Hong Yi and Liang, Mingzhao and Thongtanunam, Patanamon and Tantithamthavorn, Kla},
  title         = {Is agentic code review helpful? {Mining} developers' feedback to {CodeRabbit} reviews in the wild},
  year          = {2026},
  eprint        = {2607.03316v2},
  archivePrefix = {arXiv},
  doi           = {10.48550/arXiv.2607.03316}
}

@article{mantyla2009defects,
  author  = {M{\"a}ntyl{\"a}, Mika V. and Lassenius, Casper},
  title   = {What types of defects are really discovered in code reviews?},
  journal = {IEEE Transactions on Software Engineering},
  volume  = {35},
  number  = {3},
  pages   = {430--448},
  year    = {2009},
  doi     = {10.1109/TSE.2008.71}
}

@inproceedings{panickssery2024selfpref,
  author    = {Panickssery, Arjun and Bowman, Samuel R. and Feng, Shi},
  title     = {{LLM} evaluators recognize and favor their own generations},
  booktitle = {Advances in Neural Information Processing Systems 37 ({NeurIPS} 2024)},
  year      = {2024},
  doi       = {10.52202/079017-2197}
}

@misc{zhou2026reviewbench,
  author       = {Zhou, Michelle and Carderera de Diego, Alejandro},
  title        = {{ReviewBench}: An open benchmark for {AI} code review},
  howpublished = {The GitHub Blog},
  year         = {2026},
  month        = oct,
  url          = {https://github.com/review-bench/ReviewBench/blob/ffcb8d304c1d25caac1dee656b07ae3c1170bf04/docs/METHODOLOGY.md},
  lastaccessed = {November 18, 2026}
}

\end{document}